\documentclass[reqno,letterpaper]{amsart}

\RequirePackage[OT1]{fontenc}
\usepackage[tchypersetup]{tchdr}
\usepackage{natbib}

\numberwithin{equation}{section}

\makeatletter
\newcommand\footnoteref[1]{\protected@xdef\@thefnmark{\ref{#1}}\@footnotemark}
\makeatother
\begin{document}
\allowdisplaybreaks
\sloppy

\newcommand{\ddiam}{\overline{\eta}}
\newcommand{\doptwidth}{\nu}
\newcommand{\dnormratio}{\eta}
\newcommand{\dM}{\eta_M}

\algrenewcomment[1]{\hfill $\triangleright$ #1 }
\algnewcommand{\LineComment}[1]{\State $\triangleright$ #1 }

\title{Automated Scalable Bayesian Inference via Hilbert Coresets}
\author[T.~Campbell]{Trevor Campbell}
\address{Computer Science and Artificial Intelligence Laboratory (CSAIL)\\ Massachusetts Institute of Technology}
\urladdr{http://www.trevorcampbell.me/}
\email{tdjc@mit.edu}

\author[T.~Broderick]{Tamara Broderick}
\address{Computer Science and Artificial Intelligence Laboratory (CSAIL)\\ Massachusetts Institute of Technology}
\urladdr{http://www.tamarabroderick.com}
\email{tbroderick@csail.mit.edu}

\begin{abstract}
The automation of posterior inference in Bayesian data analysis has enabled experts and nonexperts alike 
to use more sophisticated models, engage in faster exploratory modeling and analysis, and ensure experimental reproducibility.
However, standard automated posterior inference algorithms are not tractable at the scale of massive modern datasets,
and modifications to make them so are typically model-specific, require expert tuning, and can break theoretical guarantees 
on inferential quality. Building on the Bayesian coresets framework, this work instead 
takes advantage of data redundancy to shrink the dataset itself as a preprocessing step, 
providing fully-automated, scalable Bayesian inference with theoretical guarantees.
We begin with an intuitive reformulation of Bayesian coreset construction as sparse vector sum approximation, 
and demonstrate that its automation and performance-based shortcomings arise from the use of the supremum norm.
To address these shortcomings we develop Hilbert coresets, i.e., Bayesian coresets constructed under a norm induced by an inner-product on the log-likelihood function space.
We propose two Hilbert coreset construction algorithms---one based on importance sampling, and one based on the Frank-Wolfe algorithm---along
with theoretical guarantees on approximation quality as a function of coreset size.
Since the exact computation of the proposed inner-products is model-specific, we 
automate the construction with a random finite-dimensional projection of the log-likelihood functions. 
The resulting automated coreset construction algorithm is simple to implement, 
and experiments on a variety of models with real and synthetic datasets show that 
it provides high-quality posterior approximations and a significant reduction in the computational cost of inference.
\end{abstract}

\maketitle

\section{Introduction}
Bayesian probabilistic models are a standard tool of choice in modern data analysis.
Their rich hierarchies enable intelligent sharing of information across subpopulations,
their posterior distributions provide many avenues for principled parameter estimation and uncertainty quantification, 
and they can incorporate expert knowledge through the prior.
In all but the simplest models, however, the posterior distribution is 
intractable to compute exactly, and we must resort to approximate inference algorithms. 
Markov chain Monte Carlo (MCMC) \citep[Chapters 11, 12]{Gelman13} methods are the gold standard,
due primarily to their guaranteed asymptotic exactness.
Variational Bayes (VB) \citep{Jordan99,Wainwright08} is also becoming widely used due to its tractability, 
detectable convergence, and parameter estimation performance in practice.

One of the most important recent developments in the Bayesian paradigm has been
the automation of these standard inference algorithms. Rather than having to develop, code, and tune specific instantiations
of MCMC or VB for each model, practitioners now have ``black-box'' implementations
that require only a basic specification of the model as inputs. 
For example, while standard VB requires the specification of model gradients---whose formulae
are often onerous to obtain---and an approximating family---whose rigorous selection
is an open question---ADVI \citep{Ranganath14,Kucukelbir15,Kucukelbir17} applies standard transformations to the model so that
a multivariate Gaussian approximation can be used, and computes gradients with automatic differentiation. 
The user is then left only with the much simpler task of specifying the log-likelihood and prior. 
Similarly, while Hamiltonian Monte Carlo \citep{Neal11} requires tuning a step size and path length parameter, NUTS \citep{Hoffman14}
provides a method for automatically determining reasonable values for both.
This level of automation has many benefits: it enables experts and nonexperts alike to use more sophisticated models,
it facilitates faster exploratory modeling and analysis, and helps ensure experimental reproducibility.

But as modern datasets continue to grow larger over time, it is important for inference to be not only automated,
but \emph{scalable} while retaining \emph{theoretical guarantees} on the quality of inferential results.
In this regard, the current set of available inference algorithms falls short. 
Standard MCMC algorithms may be ``exact'', but they are typically not tractable for large-scale data,
as their complexity per posterior sample scales at least linearly in the dataset size. Variational methods
on the other hand are often scalable, but posterior approximation guarantees continue to elude researchers 
in all but a few simple cases. Other scalable Bayesian inference algorithms have largely been developed by
modifying standard inference algorithms to handle distributed
or streaming data processing. Examples include subsampling and streaming methods for 
variational Bayes~\citep{Hoffman:2013,Broderick:2013b,Campbell:2015}, subsampling methods for 
MCMC~\citep{Welling:2011,Ahn:2012,Bardenet:2014,Korattikara:2014,Maclaurin:2014,Bardenet:2015}, and 
distributed ``consensus'' methods for MCMC~\citep{Scott16,Srivastava:2015,Rabinovich:2015,Entezari:2016}. 
These methods either have no guarantees on the quality of their inferential results, or require
expensive iterative access to a constant fraction of the data, but
more importantly they tend to be model-specific and require extensive expert tuning. This
makes them poor candidates for automation on the large class of models to which standard
automated inference algorithms are applicable.

An alternative approach, based on the observation that large datasets often contain redundant data,
is to modify the dataset itself such that its size is reduced while preserving its original statistical properties.
In Bayesian regression, for example, a large dataset can be compressed using random linear projection \citep{Geppert17,Ahfock17,Bardenet15}.
For a wider class of Bayesian models, one can construct a small weighted subset of the data, 
known as a \emph{Bayesian coreset}\footnote{The concept of a coreset originated in computational 
geometry and optimization \citep{Agarwal05,Feldman11,Feldman13,Bachem:2015,Lucic:2016,Bachem:2016,Feldman:2011b,Han:2016}.} \citep{Huggins16}, whose 
weighted log-likelihood approximates the full data log-likelihood. 
The coreset can then be passed to any standard (automated) inference algorithm, providing
posterior inference at a significantly reduced computational cost. 
Note that since the coresets approach is agnostic to the particular inference algorithm used,
its benefits apply to the continuing developments in both MCMC \citep{Robert18} and variational \citep{Dieng17,Li16,Liu16}
approaches.

Bayesian coresets, in contrast to other large-scale inference techniques, are simple to implement, computationally inexpensive,
and have theoretical guarantees relating coreset size to both computational complexity 
and the quality of approximation \citep{Huggins16}. 
However, their construction cannot be easily automated, as
it requires computing the
\emph{sensitivity} \citep{Langberg10} of each data point, 
a model-specific task that involves significant technical expertise. 
This approach also often necessitates a bounded parameter space to ensure bounded sensitivities,
precluding many oft-used continuous likelihoods and priors.
Further, since Bayesian coreset construction involves \iid random subsampling,
it can only reduce approximation error compared to uniform subsampling by a constant, and
cannot update its notion of importance based on what points it has already selected. 

In this work, we develop a scalable, theoretically-sound Bayesian approximation framework
with the same level of automation as ADVI and NUTS, the algorithmic simplicity 
and low computational burden of Bayesian coresets, 
and the inferential performance of hand-tuned, model-specific scalable algorithms. 
We begin with an intuitive reformulation of Bayesian coreset construction as sparse vector sum approximation,
in which the data log-likelihood functions are vectors in a vector space,
sensitivity is a weighted uniform (i.e.~supremum) norm on those vectors, and 
the construction algorithm is importance sampling. This perspective illuminates
the use of the uniform norm as the primary source of the shortcomings of Bayesian coresets.
To address these issues we develop Hilbert coresets, i.e., Bayesian coresets using a norm induced by an inner-product on the log-likelihood function space.
Our contributions include two candidate norms: one a weighted $L^2$ norm, and another based on 
the Fisher information distance \citep{Johnson04}. Given these norms, we provide an importance sampling-based coreset construction algorithm and
a more aggressive ``direction of improvement''-aware coreset construction based on the Frank--Wolfe 
algorithm \citep{Frank56,Guelat86,Jaggi13}. Our contributions include theoretical guarantees
relating the performance of both to coreset size. Since the proposed norms and inner-products cannot in general be computed
in closed-form, we automate the construction using a random finite-dimensional projection of the log-likelihood functions inspired by \citet{Rahimi07}.
We test Hilbert coresets empirically on multivariate Gaussian inference, logistic regression, Poisson regression, and von Mises-Fisher mixture
modeling with both real and synthetic data; these experiments show that Hilbert coresets 
provide high quality posterior approximations with a significant reduction in the computational cost of inference compared to standard
automated inference algorithms. 
All proofs are deferred to \cref{sec:proofs}.

\section{Background}\label{sec:background}
In the general setting of Bayesian posterior inference, 
we are given a dataset $\left(y_n\right)_{n=1}^N$ of $N$ observations,
a likelihood $p(y_n | \theta)$ for each observation given the parameter $\theta \in \Theta$,
and a prior density $\pi_0(\theta)$ on $\Theta$.
We assume throughout that the data are conditionally independent given $\theta$.
The Bayesian posterior is given by the density
\[
\pi(\theta) &\defined \frac{1}{Z}\exp(\mcL(\theta))\pi_{0}(\theta), \label{eq:bayes}
\]
where the log-likelihood $\mcL(\theta)$ and marginal likelihood $Z$ are defined by
\[
\mcL_n(\theta) \defined \log p(y_{n} \given \theta), && 
\mcL(\theta) \defined \sum_{n=1}^N\mcL_n(\theta), && 
Z \defined \int\exp(\mcL(\theta))\pi_{0}(\theta)\,\dee \theta. \label{eq:bayesdefns}
\]

In almost all cases in practice, an exact closed-form expression of $\pi$ is not available due to the difficulty
of computing $Z$, forcing the use of approximate Bayesian inference algorithms. 
While Markov chain Monte Carlo (MCMC) algorithms \citep[Chapters 11, 12]{Gelman13} 
are often preferred for their theoretical guarantees asymptotic in running time, they are typically computationally intractable for large $N$.
One way to address this is to construct a small, weighted subset of the original
dataset whose log-likelihood approximates that of the full dataset, known as a
\emph{Bayesian coreset} \citep{Huggins16}. This coreset can then be passed to a standard
MCMC algorithm. The computational savings from running MCMC on a much smaller dataset
can allow a much faster inference procedure while retaining the theoretical guarantees of MCMC.
In particular, the aim of the Bayesian coresets framework
is to find a set of nonnegative weights $w \defined (w_n)_{n=1}^N$,
a small number of which are nonzero, such that the weighted log-likelihood
\[
\mcL(w, \theta) \defined \sum_{n=1}^N w_n \mcL_n(\theta) 
\quad 
\text{satisfies}
\quad
\left|\mcL(w, \theta)- \mcL(\theta)\right| \le \epsilon\left|\mcL(\theta)\right|,  \,\, \forall \theta \in \Theta. \label{eq:likeapprox}
\]
The algorithm proposed by \citet{Huggins16} to construct a Bayesian coreset is as follows. First, compute
the \emph{sensitivity} $\sigma_n$ of each data point,
\[
\sigma_n \defined \sup_{\theta\in\Theta} \left|\frac{\mcL_n(\theta)}{\mcL(\theta)}\right|, \label{eq:sensitivity}
\]
and then subsample the dataset by taking $M$ independent draws with probability proportional to $\sigma_n$ (resulting in a coreset of size $\leq M$) via
\[
\sigma &\defined \sum_{n=1}^N \sigma_n &
(M_1, \dots, M_N) &\dist \distMulti\left(M, \left(\frac{\sigma_n}{\sigma}\right)_{n=1}^N\right) &
W_n &= \frac{\sigma}{\sigma_n} \frac{M_n}{M}.\label{eq:uniformcoreset}
\]
Since $\EE\left[W_n\right] = 1$, we have that $\EE\left[\mcL(W, \theta)\right] = \mcL(\theta)$,
and we expect that $\mcL(W, \theta) \to \mcL(\theta)$ in some sense as $M$ increases.
This is indeed the case; \citet{Braverman16,Feldman11} showed that with high probability, the coreset likelihood $\mcL(W, \theta)$ satisfies 
\cref{eq:likeapprox} with $\epsilon^2 = O\left(\frac{1}{M}\right)$, 
and \citet{Huggins16} extended this result to the case of Bayesian coresets in the setting of logistic regression.
Typically, exact computation of the sensitivities $\sigma_n$ is not tractable, so upper bounds are used instead \citep{Huggins16}.

\section{Coresets as sparse vector sum approximation}\label{sec:vectorcoresets}
This section develops an intuitive perspective of Bayesian 
coresets as sparse vector sum approximation under a uniform norm, and
draws on this perspective to uncover the limitations of the framework and avenues for extension.
Consider the vector space of functions $g : \Theta \to \reals$ with bounded uniform norm
weighted by the total log-likelihood $\mcL(\theta)$,
\[
\left\|g\right\| &\defined \sup_{\theta \in \Theta}\left|\frac{g(\theta)}{\mcL(\theta)}\right|.\label{eq:supnorm}
\]
In this space,
the data log-likelihood functions $\mcL_n(\theta)$ have vectors $\mcL_n$ with norm $\sigma_n\defined \left\|\mcL_n\right\|$ as defined in \cref{eq:sensitivity},
the total log-likelihood has vector $\mcL\defined\sum_{n=1}^N \mcL_n$, and
the coreset guarantee in \cref{eq:likeapprox} corresponds to approximation of $\mcL$ with
the vector $\mcL(w)\defined \sum_{n=1}^N w_n\mcL_n$ under the vector norm with error at most $\epsilon$, i.e.
$\left\|\mcL(w) - \mcL\right\| \leq \epsilon$. Given this formulation, we can write the problem of constructing the best coreset
of size $M$ as the minimization of approximation error subject to a constraint on the number of nonzero entries in $w$,
\[
\min_{w \in \reals^N} \quad & \left\|\mcL(w) - \mcL \right\|^2 \quad \text{s.t.} \quad  w \geq 0, \quad \sum_{n=1}^N \ind\left[w_n > 0\right] \leq M.\label{eq:fullproblem}
\]
\cref{eq:fullproblem} is a convex optimization with binary constraints, and thus is difficult to solve efficiently in general;
we are forced to use approximate methods. The uniform Bayesian coresets framework provides one such approximate method, where
$\mcL/N$ is viewed as the expectation of a uniformly random subsample of $\left(\mcL_n\right)_{n=1}^N$,
and importance sampling is used to reduce the expected error of the estimate. 
Choosing importance probabilities proportional to $\sigma_n=\left\|\mcL_n\right\|$ results in
a high-probability bound on approximation error given below in \cref{thm:supremumimportancesampling}.
The proof of \cref{thm:supremumimportancesampling} in \cref{sec:proofs}
is much simpler than similar results available in the literature~\citep{Feldman11,Braverman16,Huggins16}
due to the present vector space formulation.
\cref{thm:supremumimportancesampling} depends on two constants ($\sigma$ and $\ddiam$) that capture important aspects of the geometry  of the optimization problem: 
\[
\sigma_n &\defined \left\|\mcL_n\right\|
&
\sigma &\defined \sum_{n=1}^N \sigma_n 
&
\ddiam^2 &\defined \max_{n, m \in [N]} \left\|\frac{\mcL_n}{\sigma_n} - \frac{\mcL_m}{\sigma_m}\right\|^2,  \label{eq:sigddiamdefn}
\] 
where $[N] \defined \left\{1, 2, \dots, N\right\}$.
The quantity $\sigma \geq 0$ captures the scale of the problem; all error guarantees on $\left\|\mcL(w)-\mcL\right\|$
should be roughly linearly proportional to $\sigma$. The quantity $0 \leq\ddiam \leq 2$ captures how well-aligned the vectors $(\mcL_n)_{n=1}^N$ are,
and thus the inherent difficulty of approximating $\mcL$ with a sparse weighted subset $\mcL(w)$. For example, if all vectors are aligned then
$\ddiam = 0$, and the problem is trivial since we can achieve 0 error with a single scaled vector $\mcL_n$.
\cref{thm:supremumimportancesampling} also depends on an approximate notion of the dimension of the span
of the log-likelihood vectors $(\mcL_n)_{n=1}^N$, given by \cref{defn:dim}. Note in particular that
the approximate dimension of a set of vectors in $\reals^d$ is at most $d$, corresponding to the usual notion of dimension in this setting.
\bnumdefn\label{defn:dim}
The \emph{approximate dimension}  $\dim\left(u_n\right)_{n=1}^N$ of $N$ vectors in a normed vector space
is the minimum value of $d \in \nats$ such that all vectors $u_n$ can be approximated
using linear combinations of a set of $d$ unit vectors $(v_j)_{j=1}^d$, $\|v_j\|=1$:
\[
\forall\, n\in[N], \,\, \exists\, \alpha_n \in [-1, 1]^{d} \,\, \text{s.t.}\,\, \left\|\frac{u_n}{\|u_n\|} - \sum_{j=1}^d\alpha_{nj}v_j\right\| \leq \frac{d}{\sqrt{N}}.
\]
\enumdefn
\bnthm\label{thm:supremumimportancesampling}
Fix any $\delta \in (0, 1)$. With probability  $\geq 1-\delta$, the output of the uniform coreset construction algorithm in \cref{eq:uniformcoreset} satisfies
\[
\left\|\mcL(W)-\mcL\right\| \leq \frac{\sigma}{\sqrt{M}}\left(\frac{3}{2}\dim\left(\mcL_n\right)_{n=1}^N + \ddiam\sqrt{2\log\frac{1}{\delta}}\right).
\]
\enthm

\begin{figure}[t!]
\begin{subfigure}[t]{0.45\textwidth}
\includegraphics[width=\columnwidth, clip, trim=40 30 40 8]{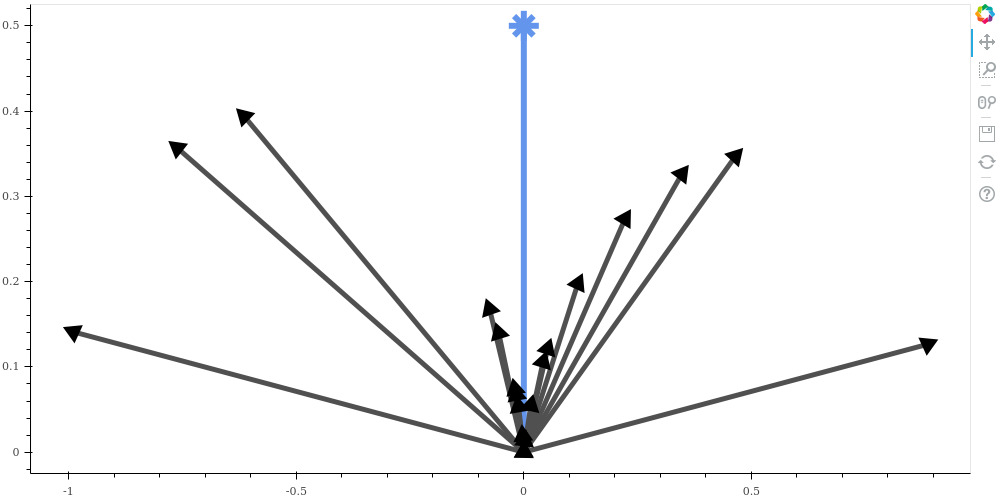}
\caption{}\label{fig:intuition_vec}
\end{subfigure}
\begin{subfigure}[t]{0.45\textwidth}
\includegraphics[width=\columnwidth, clip, trim=40 30 40 8]{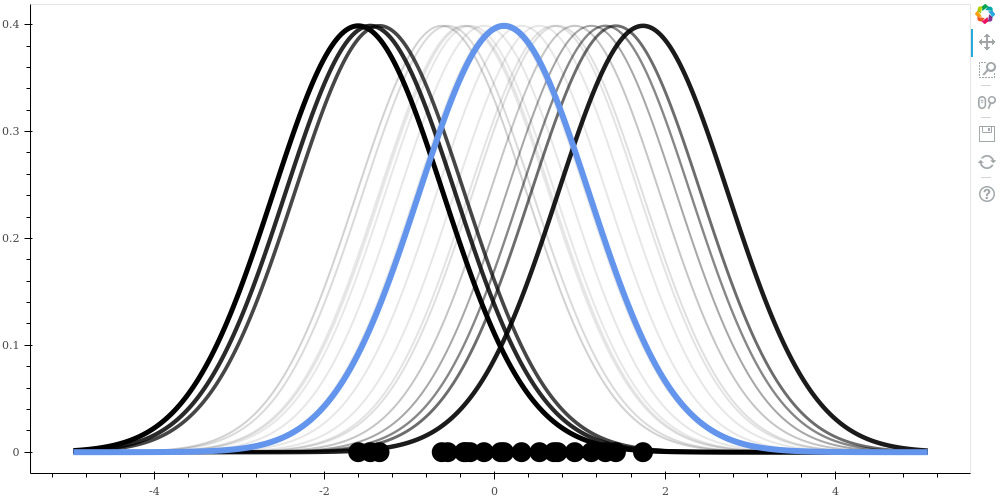}
\caption{}\label{fig:intuition_gaus}
\end{subfigure}
\caption{(\ref{fig:intuition_vec}): The sparse vector approximation problem, depicting the sum $\mcL$ in blue and the vectors $\mcL_n$ in grey.
(\ref{fig:intuition_gaus}): Uniform Bayesian coresets behavior on the simple exercise of learning a Gaussian mean.
Depicted are data likelihoods in black, scaled posterior density in blue, and data as black scatter points. 
The sensitivity of each datapoint is indicated by the line thickness of the likelihood---a thicker, darker
line denotes higher sensitivity, while thinner, lighter lines denote lower sensitivity. }\label{fig:intuition}
\end{figure}

The analysis, discussion, and algorithms presented to this point are independent of the particular choice of norm given in \cref{eq:supnorm};
one might wonder if the uniform norm used above is the best choice, or if there is another norm more suited to Bayesian inference in some way.
For instance, the supremum in \cref{eq:supnorm} can diverge in an unbounded or infinite-dimensional parameter space $\Theta$, requiring an artificial restriction placed on the space
\citep{Huggins16}.
This precludes the application to the many common models and priors that have unbounded parameter spaces, even
logistic regression with full support $\Theta = \reals^d$.
The optimization objective function in \cref{eq:supnorm} is also typically nonconvex, and finding (or bounding) the optimum 
is a model-specific task that is not easily automated. 

Perhaps most importantly, the uniform norm lacks a sense of ``directionality'' as it does not correspond to an inner-product.
This implies that the bound in \cref{thm:supremumimportancesampling} does not scale properly with
the alignment of vectors (note how the error does not approach 0 as $\ddiam\to 0$)
and that it depends on the approximate dimension (which may be hard to compute).
Moreover, the lack of directionality makes the coreset construction algorithm behave counterintuitively and limits its performance in a fundamental way. 
\cref{fig:intuition_vec} provides a pictorial representation of this limitation. Recall that the goal of coreset construction is to find a \emph{sparse} weighted subset 
of the vectors $\left(\mcL_n\right)_{n=1}^N$ (grey) that approximates $\mcL$ (blue).
In this example, there are vectors which, when scaled, could individually nearly perfectly replicate $\mcL$. But
the importance sampling algorithm in \cref{eq:uniformcoreset} will instead tend to sample those vectors with large norm that are pointed away from $\mcL$,
requiring a much larger coreset to achieve the same approximation error.
This is a consequence of the lack of directionality of the uniform norm; it has no concept of the alignment
of certain vectors with $\mcL$, and is forced to mitigate worst-case error by sampling those vectors
with large norm. 
\cref{fig:intuition_gaus} shows the result of this behavior in a 1D Gaussian inference problem.
In this figure, the likelihood functions of the data are depicted in black, with their uniform norm (or sensitivity) indicated
by thickness and opacity. The posterior distribution is displayed in blue, with its log-density scaled by $1/N$ for clarity. 
The importance sampling algorithm in \cref{eq:uniformcoreset} will tend to sample those data that are \emph{far away} from the posterior
mean, with likelihoods that are different than the scaled posterior, despite the fact that there are data close to the mean whose likelihoods are near-perfect
approximations of the scaled posterior.
Using the intuition from \cref{fig:intuition}, it is not difficult to construct examples where the expected error of importance sampling is arbitrarily 
worse than the error of the optimal coreset of size $M$.

\section{Hilbert coresets}\label{sec:hilbert}
It is clear that a notion of directionality of the vectors $\left(\mcL_n\right)_{n=1}^N$ 
is key to developing both efficient, intuitive coreset construction algorithms and theory that correctly reflects problem difficulty.
Therefore, in this section we develop methods for constructing Bayesian coresets in a Hilbert space (\emph{Hilbert coresets}), i.e., using a norm corresponding to an inner product.
The notion of directionality granted by the inner product provides two major advantages over uniform coresets: 
coreset points can be chosen intelligently based on the residual posterior approximation error vector; and
theoretical guarantees on approximation quality can directly incorporate the difficulty of the approximation problem
via the alignment of log-likelihood vectors. We provide two coreset construction algorithms which take advantage of these benefits.
The first method, developed in \cref{sec:hilbertis}, is based on viewing $\mcL/N$ as the expectation of a uniformly random subsample of
$\left(\mcL_n\right)_{n=1}^N$, and then using importance sampling to reduce the expected error of the estimate.
The second method, developed in \cref{sec:hilbertfw}, is based on viewing the cardinality-unconstrained version of \cref{eq:fullproblem} as a quadratic optimization over an
appropriately-chosen polytope, and then using the Frank--Wolfe algorithm \cite{Frank56,Guelat86,Jaggi13} to compute a sparse approximation to the optimum.
Theoretical guarantees on posterior approximation error are provided for both.
In \cref{sec:hilbertdistributed}, we develop streaming/distributed extensions of these methods and provide similar approximation guarantees.
Note that this section treats the general case of Bayesian coreset construction with a Hilbert space norm; 
the selection of a particular norm and its automated computation is left to \cref{sec:bayeshilbert}.

\subsection{Coreset construction via importance sampling}\label{sec:hilbertis}
Taking inspiration from the uniform Bayesian coreset construction algorithm,
the first Hilbert coreset construction method, \cref{alg:importancesampling}, 
involves \iid sampling from the vectors $\left(\mcL_n\right)_{n=1}^N$ with probabilities
$\left(p_n\right)_{n=1}^N$ and reweighting the subsample. 
In contrast to the case of the weighted uniform norm in \cref{eq:supnorm}, 
the choice $p_n \propto \sigma_n$ exactly minimizes the expected squared coreset error 
under a Hilbert norm (see \cref{eq:optimalsamplingprobs} in \cref{sec:proofs}), yielding
\[
\EE\left[\left\|\mcL(W) - \mcL\right\|^2\right] &= \frac{\sigma^2\eta^2}{M} &
\dnormratio^2 &\defined 1 - \frac{\|\mcL\|^2}{\sigma^2},\label{eq:dnormratiodefn}
\]
where $0\leq \dnormratio \leq 1$, similar to $\ddiam$, captures how well-aligned the vectors $\left(\mcL_n\right)_{n=1}^N$ are.
However, in a Hilbert space $\dnormratio$ is a tighter constant: $\dnormratio \leq \ddiam/\sqrt{2}$ by \cref{lem:dnormratiovsddiam}.
\cref{thm:importancesampling}, whose proof in \cref{sec:proofs} relies on standard martingale concentration inequalities,
 provides a high-probability guarantee  on the quality of the output approximation.
This result depends on $\ddiam$ from \cref{eq:sigddiamdefn} and $\dnormratio$ from \cref{eq:dnormratiodefn}.
\begin{algorithm}[t!]
\caption{IS: Hilbert coresets via importance sampling}\label{alg:importancesampling}
\begin{algorithmic}
\Require $(\mcL_n)_{n=1}^N$, $M$, $\left\|\cdot\right\|$
\State $\forall n \in [N] \, \, \sigma_n \gets \|\mcL_n\|$, and $\sigma \gets \sum_{n=1}^N \sigma_n$ \Comment{compute norms}
\State $\left(M_1, \dots, M_N\right) \gets \distMulti\left(M, \left(\frac{\sigma_n}{\sigma}\right)_{n=1}^N\right)$ \Comment{subsample the data}
\State $W_n \gets \frac{\sigma}{\sigma_n} \frac{M_n}{M}$ for $n\in[N]$ \Comment{reweight the subsample}
\State \Return $W$
\end{algorithmic}
\end{algorithm}
\bnthm\label{thm:importancesampling}
Fix any $\delta \in (0, 1)$. With probability  $\geq 1-\delta$,
the output $W$ of \cref{alg:importancesampling} satisfies
\[
\left\|\mcL(W) - \mcL\right\| & \leq \frac{\sigma}{\sqrt{M}}\left(\dnormratio + 
 \dM \sqrt{2\log\frac{1}{\delta}}\right)
\]
where
\[
 \dM &\defined \min\left( \ddiam,  \dnormratio\sqrt{\frac{2M\dnormratio^2}{\ddiam^2\log\frac{1}{\delta}}}H^{-1}\!\!\left(\frac{\ddiam^2\log\frac{1}{\delta}}{2M \dnormratio^2}\right)\right)\\
H(y) &\defined (1+y)\log(1+y)-y.
\]
\enthm
In contrast to \cref{thm:supremumimportancesampling}, \cref{thm:importancesampling} takes advantage of the inner product to incorporate a notion of problem difficulty into the bound. 
For example, since $H(y) \sim y^2$ as $y \to 0$, we have that $H^{-1}(y) \sim \sqrt{y}$ and so $\lim_{y\to 0} \sqrt{y^{-1}} H^{-1}(y)  = 1$. 
Combined with the fact that $\dnormratio \leq \ddiam$, we have $\lim_{M\to\infty} \dM  = \dnormratio$, and so 
the bound in \cref{thm:importancesampling} is asymptotically equivalent to $\frac{\sigma\dnormratio}{\sqrt{M}}\left(1 +  \sqrt{2\log\frac{1}{\delta}}\right)$ as $M\to\infty$. 
Given that importance sampling can only improve convergence over uniformly random subsampling by a constant, this constant reduction is significant.
Note that \cref{thm:importancesampling}, in conjunction with the fact that $\dnormratio \leq \ddiam$, 
immediately implies the simpler result in \cref{cor:importancesampling}.
\bncor\label{cor:importancesampling}
Fix any $\delta \in (0, 1)$. With probability  $\geq 1-\delta$,
the output $W$ of \cref{alg:importancesampling} satisfies
\[
\left\|\mcL(W) - \mcL\right\| & \leq \frac{\sigma\ddiam}{\sqrt{M}}\left(1 +  \sqrt{2\log\frac{1}{\delta}}\right).
\]
\encor

\subsection{Coreset construction via Frank--Wolfe}\label{sec:hilbertfw}
The major advantages of \cref{alg:importancesampling} are its simplicity and sole requirement of computing the norms $(\|\mcL_n\|)_{n=1}^N$. 
Like the original uniform Bayesian coreset algorithm in \cref{eq:uniformcoreset}, however,
it does not take into account the residual error in the coreset approximation in order to choose new samples intelligently.
The second Hilbert coreset construction method, \cref{alg:frankwolfe}, takes advantage of the directionality of the Hilbert norm
to incrementally build the coreset by selecting vectors aligned with the ``direction of greatest improvement.''

The development of \cref{alg:frankwolfe} involves two major steps. First, we replace the cardinality constraint on $w$ in \cref{eq:fullproblem}
with a polytope constraint:
\[
\min_{w \in \reals^N} \quad & (w-1)^TK(w-1) \quad
\text{s.t.} \quad w\geq0, \quad \sum_{n=1}^N \sigma_n w_n = \sigma,\label{eq:modifiedproblem}
\]
where $K\in\reals^{N\times N}$ is a kernel matrix defined by $K_{ij} \defined \left<\mcL_i, \mcL_j\right>$,
and we take advantage of the Hilbert norm to rewrite $\left\|\mcL(w)-\mcL\right\|^2 = (w-1)^TK(w-1)$.
The polytope is designed to contain the point $w = 1 \defined \left[1, 1, \dots, 1\right]^T\in\reals^N$---which is optimal with cost 0 since $\mcL(1) = \mcL$---and 
have vertices $\frac{\sigma}{\sigma_n} 1_n$ for $n\in[N]$, where $1_n$ is the $n^\text{th}$ coordinate unit vector.
Next, taking inspiration from the large-scale optimization literature \cite{Frank56,Guelat86,Jaggi13,LacosteJulien15,Clarkson10,Reddi16,Balasubramanian18,Hazan16}, we solve the convex optimization in \cref{eq:modifiedproblem} using the Frank--Wolfe algorithm \citep{Frank56}.
Frank--Wolfe is an iterative algorithm for solving convex optimization problems of the form $\min_{x\in\mcD} g(x)$, where
each iteration has three steps: 1) given the $t^\text{th}$ iterate $x_t$, we first find a search direction $d_t = s_t - x_t$ by solving the 
linear program $s_t = \argmin_{s\in\mcD} \grad f(x_t)^T s$; 
2) we find a step size by solving the 1-dimensional optimization $\gamma_{t} = \argmin_{\gamma\in[0,1]} f(x_t + \gamma d_t)$;
and 
3) we update $x_{t+1} \gets x_t + \gamma_td_t$.
In \cref{eq:modifiedproblem}, we are optimizing a convex objective over a polytope, so
the linear optimization can be solved by searching over all vertices of the polytope.
And since we designed the polytope such that its vertices each have a single nonzero component, the algorithm 
adds at most a single data point to the coreset
at each iteration; after initialization followed by $M-1$ iterations, this produces a coreset of size $\leq M$. 

\begin{algorithm}[t!]
\caption{FW: Hilbert coresets via Frank--Wolfe}\label{alg:frankwolfe}
\begin{algorithmic}
\Require $(\mcL_n)_{n=1}^N$, $M$, $\left<\cdot, \cdot\right>$
\State $\forall n \in [N] \, \, \sigma_n \gets \sqrt{\left<\mcL_n,\mcL_n\right>}$, and $\sigma \gets \sum_{n=1}^N \sigma_n$ \Comment{compute norms}
\State $f \gets \argmax_{n\in[N]} \left<\mcL, \frac{1}{\sigma_n}\mcL_n\right>$ \Comment{greedy initial vertex $f$ selection}
\State $w \gets \frac{\sigma}{\sigma_f} 1_f$ \Comment{initialize $w$ with full weight on $f$}
\For{$t \in \{1, \dots, M-1\}$}
\State $f \gets \argmax_{n \in [N]} \left<\mcL - \mcL(w),  \frac{1}{\sigma_n}\mcL_n \right>$ \Comment{find the FW vertex index $f$}
\State $\gamma \gets \frac{\left< \frac{\sigma}{\sigma_f} \mcL_f - \mcL(w), \mcL - \mcL(w)\right>}{\left< \frac{\sigma}{\sigma_f} \mcL_f - \mcL(w), \frac{\sigma}{\sigma_f}\mcL_f - \mcL(w)\right>}$ \Comment{closed-form line search for step size $\gamma$}
\State $w \gets (1-\gamma)w + \gamma \frac{\sigma}{\sigma_f}1_f$ \Comment{add/reweight data point $f$ in coreset}
\EndFor
\State \Return $w$
\end{algorithmic}
\end{algorithm}

We initialize $w_0$ to the vertex most aligned with $\mcL$, i.e.
\[
w_0 &= \frac{\sigma}{\sigma_{f_0}}1_{f_0}
\quad \text{where} \quad
f_0 = \argmax_{n \in [N]} \left<\mcL, \frac{1}{\sigma_n}\mcL_n\right>. \label{eq:fwinit}
\]
Let $w_t$ be the iterate at step $t$. The gradient of the cost is $2 K\left(w_t - 1\right)$, 
and we are solving a convex optimization on a polytope, so the Frank--Wolfe direction may be computed
by searching over its vertices: 
\[
d_t \defined \frac{\sigma}{\sigma_{f_t}} 1_{f_t} - w_t
\quad \text{where}\quad
f_t 
 = \argmax_{n\in[N]}\left<\mcL - \mcL(w_t), \frac{1}{\sigma_n}\mcL_n\right>. \label{eq:fwdir}
\]
The Frank--Wolfe algorithm applied to \cref{eq:modifiedproblem} thus corresponds to a simple greedy approach in which 
we select the vector $\mcL_{f_t}$ with direction most aligned with the residual error $\mcL-\mcL(w_t)$.
We perform line search to update $w_{t+1} = w_t + \gamma d_t$ for some $\gamma \in [0, 1]$.
 Since the objective is quadratic, the exact solution for unconstrained line search is available in closed form per \cref{eq:linesearch}; \cref{lem:linesearch} shows
that this is actually the solution to constrained line search in $\gamma\in[0, 1]$, ensuring that $w_{t+1}$ remains feasible.
\[
w_{t+1} &= w_t + \gamma_td_t  
\quad \text{where} \quad 
\gamma_t = \frac{\left< \frac{\sigma}{\sigma_{f_t}} \mcL_{f_t} - \mcL(w_t), \mcL - \mcL(w_t)\right>}{\left< \frac{\sigma}{\sigma_{f_t}} \mcL_{f_t} - \mcL(w_t), \frac{\sigma}{\sigma_{f_t}} \mcL_{f_t} - \mcL(w_t)\right>}.\label{eq:linesearch}
\]
\bnlem\label{lem:linesearch}
For all $t\in\nats$, $\gamma_t\in[0, 1]$.
\enlem

\cref{thm:frankwolfe} below provides a guarantee on the quality of the approximation output by \cref{alg:frankwolfe} 
using the combination of the initialization in \cref{eq:fwinit} and 
exact line search in \cref{eq:linesearch}.
This result depends on the constants $\ddiam$ from \cref{eq:sigddiamdefn}, $\dnormratio$ from \cref{eq:dnormratiodefn},
and $\doptwidth$, defined by
\[
\doptwidth^2 \defined 1-\frac{r^2}{\sigma^2\ddiam^2}, \label{eq:doptwidth}
\]
where $r$ is the distance from $\mcL$ to the nearest boundary of the convex hull of $\left\{\sigma\mcL_n/\sigma_n\right\}_{n=1}^N$.
Since $\mcL$ is in the relative interior of this convex hull by \cref{lem:relint}, we are guaranteed that $\doptwidth < 1$. 
The proof of \cref{thm:frankwolfe} in \cref{sec:proofs} relies
on a technique from \citet[Theorem 2]{Guelat86} and a novel bound on the logistic equation.
\bnthm\label{thm:frankwolfe}
The output $w$ of \cref{alg:frankwolfe} satisfies
\[
\left\|\mcL(w) - \mcL\right\| &\leq \frac{\sigma\dnormratio\ddiam\doptwidth}{\sqrt{\ddiam^2\doptwidth^{-2(M-2)} + \dnormratio^2(M-1)}} \leq \frac{\sigma\ddiam}{\sqrt{M}}.
\]
\enthm
In contrast to previous convergence analyses of Frank--Wolfe optimization, \cref{thm:frankwolfe} exploits the quadratic objective
and exact line search to capture both the logarithmic $\nicefrac{1}{\sqrt{M}}$ convergence rate for small values of $M$, and the linear $\doptwidth^{M}$ rate
for large $M$.  Alternatively, one can remove the computational cost of computing the exact line search via \cref{eq:linesearch} by simply setting
$\gamma_t = \frac{2}{3t+4}$. In this case, \cref{thm:frankwolfe} is replaced with the weaker result (see the note at 
the end of the proof of \cref{thm:frankwolfe} in \cref{sec:proofs})
\[
\left\|\mcL(w) - \mcL\right\| &\leq \frac{2\sigma\ddiam}{\sqrt{3M+1}}.
\]

\subsection{Distributed coreset construction}\label{sec:hilbertdistributed}
An advantage of Hilbert coresets---and coresets in general---is that they apply to streaming and distributed
data with little modification, and retain their theoretical guarantees. In particular,
if the dataset $\left(y_n\right)_{n=1}^N$ is distributed (not necessarily evenly) among $C$ processors, 
and either \cref{alg:importancesampling} or \cref{alg:frankwolfe} are run for $M$ iterations on each processor,
the resulting merged coreset of size $\leq MC$ has an error guarantee given by \cref{cor:distributedis} or \cref{cor:distributedfw}.
Note that the weights from each distributed coreset are not modified when merging.
These results both follow from \cref{thm:importancesampling,thm:frankwolfe} with straightforward usage of
the triangle inequality, the union bound, and the fact that $\ddiam$ for each subset is bounded above by $\ddiam$ for the full dataset.
\bncor\label{cor:distributedis}
Fix any $\delta\in(0, 1)$. With probability $\geq 1-\delta$,
 the coreset constructed by running \cref{alg:importancesampling} on $C$ nodes 
and merging the result satisfies
\[
\left\|\mcL(w)-\mcL\right\| &\leq \frac{\sigma\ddiam}{\sqrt{M}}\left(1+\sqrt{2\log\frac{C}{\delta}}\right).
\]
\encor
\bncor\label{cor:distributedfw}
The coreset constructed by running \cref{alg:frankwolfe} on $C$ nodes and merging the results satisfies
\[
\left\|\mcL(w) - \mcL\right\| &\leq \frac{\sigma\ddiam}{\sqrt{M}}.
\]
\encor

\section{Norms and random projection}\label{sec:bayeshilbert}
The algorithms and theory in \cref{sec:hilbert} address the scalability and performance of Bayesian coreset construction, 
but are specified for an arbitrary Hilbert norm; 
it remains to choose a norm suitable for automated Bayesian posterior approximation. There are
two main desiderata for such a norm: it should be a good indicator of posterior discrepancy,
and it should be efficiently computable or approximable in such a way that makes 
\cref{alg:importancesampling,alg:frankwolfe} efficient for large $N$, i.e., 
$O(N)$ time complexity. 
To address the desideratum that the norm is an indicator of posterior discrepancy, we propose
the use of one of two Hilbert norms. It should, however, be noted that these are simply reasonable suggestions,
and other Hilbert norms could certainly be used in the algorithms set out in \cref{sec:hilbert}. 
The first candidate is the expectation of the 
squared 2-norm difference between the log-likelihood gradients under a weighting distribution $\hat\pi$,
\[
\left\|\mcL(w)-\mcL\right\|_{\hat\pi, F}^2 &\defined\EE_{\hat \pi} \left[\left\| \grad\mcL(\theta) - \grad \mcL(w,\theta)\right\|_2^2\right], \label{eq:normF}
\]
where the weighting distribution $\hat \pi$ has the same support as the posterior $\pi$. 
This norm is a weighted version of the Fisher information distance \citep{Johnson04}.
The inner product induced by this norm is defined by
\[
\left<\mcL_n, \mcL_m\right>_{\hat \pi, F} &\defined \EE_{\hat \pi} \left[\grad \mcL_n(\theta)^T \grad\mcL_m(\theta)\right]. \label{eq:innerprodF}
\]
Although this norm has a connection to previously known discrepancies between probability distributions, it does require
that the likelihoods are differentiable. One could instead employ a simple weighted $L^2$ norm on the log-likelihoods, given by
\[
\left\|\mcL(w)-\mcL\right\|_{\hat\pi, 2} &\defined \EE_{\hat\pi}\left[\left(\mcL(\theta)  - \mcL(w,\theta)\right)^2\right] \label{eq:norm2}
\]
with induced inner product
\[
\left<\mcL_n, \mcL_m\right>_{\hat\pi, 2} &\defined \EE_{\hat\pi} \left[\mcL_n(\theta)\mcL_m(\theta)\right]. \label{eq:innerprod2}
\]
In both cases, the weighting distribution $\hat\pi$ would ideally be chosen equal to $\pi$ to emphasize discrepancies that are in regions of high posterior mass.
Though we do not have access to the true posterior without incurring significant computational cost,
there are many practical options for setting $\hat \pi$, including: the Laplace approximation \citep[Section 4.4]{Bishop06}, a posterior based on approximate sufficient statistics \citep{Huggins17}, a discrete distribution based on samples from an MCMC algorithm run on a small random subsample of the data,
the prior, independent posterior conditionals (see \cref{eq:vmfsamplewtsctrs} in \cref{sec:expt_methods}), or any other reasonable method for finding a low-cost posterior approximation. 
This requirement of a low-cost approximation is not unusual, as previous coreset formulations 
have required similar preprocessing to compute sensitivities, e.g., a $k$-clustering of the data \citep{Huggins16,Lucic:2016,Braverman16}.
We leave the general purpose selection of a weighting function $\hat\pi$ for future work. 

The two suggested norms $\left\|\cdot\right\|_{\hat\pi, 2/F}$ often do not admit exact closed-form evaluation due to the intractable expectations in \cref{eq:innerprodF,eq:innerprod2}.
Even if closed-form expressions are available, \cref{alg:frankwolfe} is computationally intractable 
when we only have access to inner products between pairs of individual log-likelihoods $\mcL_n$, $\mcL_m$, since 
obtaining the Frank--Wolfe direction involves the $O(N^2)$ computation $\argmax_{n\in[N]}\sum_{m=1}^N\left<\mcL_m, \mcL_n/\sigma_n\right>$.
Further, the analytic evaluation of expectations is a model- (and $\hat\pi$-) specific procedure that cannot be easily automated.
To both address these issues and automate Hilbert coreset construction,
 we use \emph{random features} \citep{Rahimi07}, i.e.~a random projection of the vectors $(\mcL_n)_{n=1}^N$ into a 
finite-dimensional vector space using samples from $\hat\pi$. 
For the weighted Fisher information inner product in \cref{eq:innerprodF}, we approximate $\left<\mcL_n,\mcL_m\right>_{\hat\pi, F}$ with an unbiased estimate given by
\[
(d_j)_{j=1}^J &\distiid \distUnif(\left\{1, \dots, D\right\}) \quad (\mu_j)_{j=1}^J \distiid \hat\pi\\
\left<\mcL_n, \mcL_m\right>_{\hat\pi, F} &\approx \frac{D}{J}\sum_{j=1}^J(\grad\mcL_n(\mu_j))_{d_j}(\grad\mcL_m(\mu_j))_{d_j},
\]
where subscripts indicate the selection of a component of a vector. If we define the $J$-dimensional vector
\[
\hat\mcL_n &\defined \sqrt{\frac{D}{J}}\left[(\grad\mcL_n(\mu_1))_{d_1}, (\grad\mcL_n(\mu_2))_{d_2}, \dots, (\grad\mcL_n(\mu_J))_{d_J}\right]^T,
\label{eq:randomfeaturevecF}
\]
we have that for all $n, m \in [N]$,
\[
\left<\mcL_n, \mcL_m\right>_{\hat\pi, F} &\approx \hat\mcL_n^T\hat\mcL_m.
\]
Therefore, for $n\in[N]$, $\hat\mcL_n$ serves as a random finite-dimensional projection of $\mcL_n$ that can be used in \cref{alg:importancesampling,alg:frankwolfe}.
Likewise, for the weighted $L^2$ inner product in \cref{eq:innerprod2}, we have
\[
(\mu_j)_{j=1}^J &\distiid \hat\pi \quad \left<\mcL_n, \mcL_m\right>_{\hat\pi, 2} \approx \frac{1}{J}\sum_{j=1}^J\mcL_n(\mu_j)\mcL_m(\mu_j),
\]
and so defining
\[
\hat\mcL_n &\defined \sqrt{\frac{1}{J}}\left[\mcL_n(\mu_1), \mcL_n(\mu_2), \dots, \mcL_n(\mu_J)\right]^T,
\label{eq:randomfeaturevec2}
\]
we have that for all $n, m \in [N]$,
\[
\left<\mcL_n, \mcL_m\right>_{\hat\pi, 2} &\approx \hat\mcL_n^T\hat\mcL_m,
\]
and once again $\forall n\in[N]$, $\hat\mcL_n$ serves as a finite-dimensional approximation of $\mcL_n$.
\begin{algorithm}[t!]
\caption{Bayesian Hilbert coresets with random projection}\label{alg:bayesfrankwolfe}
\begin{algorithmic}
\Require $(\mcL_n)_{n=1}^N$, $\hat\pi$, $M$, $J$
\LineComment{sample feature points and gradient dimension indices}
\State $(\mu_j)_{j=1}^J \distiid \hat\pi$, $\, \,$ $(d_j)_{j=1}^J \distiid \distUnif(\{1, \dots, D\})$ 
\LineComment{construct the random projection using one of the norms from \cref{eq:randomfeaturevecF,eq:randomfeaturevec2}}
\State If $\left\|\cdot\right\|_{\hat\pi, F}$:  $\forall n\in[N]$, $v_n \gets \sqrt{D/J}\left[(\grad\mcL_n(\mu_1))_{d_1}, \dots, (\grad\mcL_n(\mu_J))_{d_J}\right]^T$
\State If $\left\|\cdot\right\|_{\hat\pi, 2}$: $\forall n\in[N]$, $v_n \gets \sqrt{1/J}\left[\mcL_n(\mu_1), \dots, \mcL_n(\mu_J)\right]^T$
\LineComment{return the coreset constructed using random feature vectors}
\State \Return \text{FW}$\left(\left(v_n\right)_{n=1}^N, M, \left(\cdot\right)^T\!\left(\cdot\right)\right)$ or \text{IS}$\left(\left(v_n\right)_{n=1}^N, M, \left\|\cdot\right\|_2\right)$
\end{algorithmic}
\end{algorithm}
The construction of the random projections is both easily automated and enables the efficient computation of inner products with vector sums.
For example, to obtain the Frank--Wolfe direction, rather than computing $\argmax_{n\in[N]}\sum_{m=1}^N\left<\mcL_m, \mcL_n/\sigma_n\right>$,
we can simply compute $\hat\mcL = \sum_{n=1}^N \hat\mcL_n$ in $O(NJ)$ time once at the start of the algorithm and then 
$\argmax_{n\in[N]}\frac{1}{\sigma_n}\hat\mcL^T\hat\mcL_n$ in $O(NJ)$ time at each iteration.
Further, since $\hat\mcL_n^T\hat\mcL_m$ is an unbiased estimate of $\left<\mcL_n, \mcL_m\right>_{\hat\pi, 2/F}$,
we expect the error of the approximation to decrease with the random finite projection dimension $J$. \cref{thm:bayesfw} (below), whose proof may be found 
in \cref{sec:proofs}, shows that under reasonable conditions 
this is indeed the case: the difference between the true output error and random projection output error shrinks as $J$ increases.
Note that \cref{thm:bayesfw} is quite loose, due to its reliance on a max-norm quadratic form upper bound.
\bnumdefn{\citep[p.~24]{Boucheron13}}
A random variable $X$ is \emph{sub-Gaussian with constant} $\xi^2$ if 
\[
\forall \lambda \in \reals, \quad \EE\left[e^{\lambda X}\right] \leq e^{\frac{\lambda^2 \xi^2}{2}}.
\]
\enumdefn
\bnthm\label{thm:bayesfw}
Let $\mu \dist \hat\pi$, $d \dist \distUnif(\{1, \dots, D\})$, and
suppose $D \grad \mcL_n(\mu)_d \grad\mcL_m(\mu)_d$ (given $\left\|\cdot\right\|_{\hat\pi, F}$) or $\mcL_n(\mu)\mcL_m(\mu)$ (given $\left\|\cdot\right\|_{\hat\pi, 2}$)
is sub-Gaussian with constant $\xi^2$. 
Fix any $\delta\in(0, 1)$.
With probability $\geq 1-\delta$, the output of \cref{alg:bayesfrankwolfe} satisfies
\[
\left\|\mcL - \mcL(w)\right\|_{\hat\pi, 2/F}^2 &\leq \|\hat\mcL - \hat\mcL(w)\|_2^2 + \|w-1\|_1^2\sqrt{\frac{2\xi^2}{J}\log\frac{2N^2}{\delta}}.
\]
\enthm

\section{Synthetic evaluation}\label{sec:comparison}
In this section, we compare Hilbert coresets to uniform coresets and uniformly random subsampling
in a synthetic setting where expressions for the exact and coreset posteriors, along with the KL-divergence
between them, are available in closed-form. In particular, the methods are used to
perform posterior inference for the unknown mean $\mu \dist\distNorm(\mu_0, I)$ of a 2-dimensional multivariate normal distribution 
with known covariance $I$ from a collection of $N=1,000$ \iid observations $(y_n)_{n=1}^N$:
\[
\mu \dist \distNorm(\mu_0, I) \quad \left(y_n\right)_{n=1}^N \given \mu \distiid \distNorm(\mu, I).
\]

\subsection*{Methods}
We ran 1000 trials of data generation followed by uniformly random subsampling (\texttt{Rand}), uniform coresets (\texttt{Unif}),
Hilbert importance sampling (\texttt{IS}), and Hilbert Frank--Wolfe (\texttt{FW}). For the two Hilbert coreset constructions, 
we used the weighted Fisher information distance in \cref{eq:normF}. 
In this simple setting,  the exact posterior distribution is a multivariate Gaussian with mean $\mu_\pi$ and covariance $\Sigma_\pi$ given by
\[
\mu \given \left(y_n\right)_{n=1}^N &\dist \distNorm(\mu_\pi, \Sigma_\pi) &
\Sigma_\pi &= \frac{1}{1+N}I &
\mu_\pi &= \Sigma_\pi\left(\mu_0 + \sum_{n=1}^N y_n\right). 
\]
For uniform coreset construction, we subsampled the dataset as per \cref{eq:uniformcoreset}, 
where the sensitivity of $y_n$ (see \cref{sec:normalderiv} for the derivation) is given by
\[
\sigma_n  &= \frac{1}{N}\left(1 + \frac{\left(y_n-\by\right)^T\!\!\left(y_n-\by\right)}{\frac{1}{N}\sum_{m=1}^Ny_m^Ty_m - \by^T\by}\right), \quad \by\defined\frac{1}{N}\sum_{n=1}^Ny_n.\label{eq:gaussiansensitivity}
\]
This resulted in a multivariate Gaussian uniform coreset posterior approximation with mean $\hat\mu_{\pi}$ and covariance $\hat\Sigma_{\pi}$ given by
\[
\hat\Sigma_{\pi} &= \frac{1}{1+\sum_{n=1}^N W_n} I & \hat\mu_{\pi} &= \hat\Sigma_{\pi}\left(\mu_0 + \sum_{n=1}^NW_ny_n\right). \label{eq:gaussuniformcoresetpost}
\]
Generating a uniformly random subsample posterior approximation involved a similar technique, instead using probabilities $\frac{1}{N}$ for all $n\in[N]$.
For the Hilbert coreset algorithms, we used the true posterior as the weighting distribution, i.e., $\hat\pi = \pi$. This
was chosen to illustrate the ideal case in which the true posterior $\pi$ is well-approximated
by the weighting distribution $\hat\pi$. Given this choice, the Fisher information distance inner product is available in closed-form:
\[
\left<\mcL_n, \mcL_m\right>_{\pi, F} &= \frac{2}{1+N} + \left(\mu_\pi-y_n\right)^T \left(\mu_\pi - y_m\right).\label{eq:gaussianhilbertinnerprodF}
\]
Note that the norm $\|\mcL_n\|_{\pi, F}$ implied by \cref{eq:gaussianhilbertinnerprodF} and the uniform sensitivity from \cref{eq:gaussiansensitivity} are
functionally very similar; both scale with the squared distance from $y_n$ to an estimate of $\mu$.
Since all the approximate posteriors are multivariate Gaussians of the form \cref{eq:gaussuniformcoresetpost}---with weights $W_n$ differing depending on the construction algorithm---we
were able to evaluate posterior approximation quality exactly using the KL-divergence from the approximate coreset posterior $\tilde\pi$ to $\pi$,
given by
\[
\kl{\pi}{\tilde\pi} &= \frac{1}{2}\left\{\tr\left(\Sigma_{\tilde\pi}^{-1}\Sigma_{\pi}\right) + (\mu_{\tilde\pi}-\mu_\pi)^T\Sigma_{\tilde\pi}^{-1}(\mu_{\tilde\pi}-\mu_\pi) - 2 + \log\frac{\left|\Sigma_{\tilde\pi}\right|}{\left|\Sigma_\pi\right|}\right\}.
\]

\begin{figure}[t!]
\begin{subfigure}[t]{0.45\textwidth}
\includegraphics[width=.32\columnwidth, clip, trim=100 100 80 80]{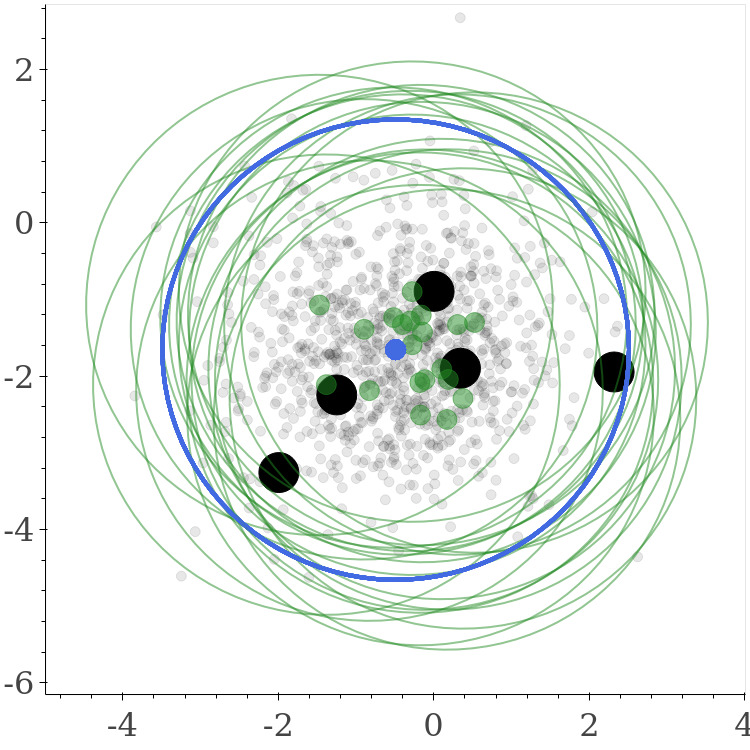}
\includegraphics[width=.32\columnwidth, clip, trim=100 100 80 80]{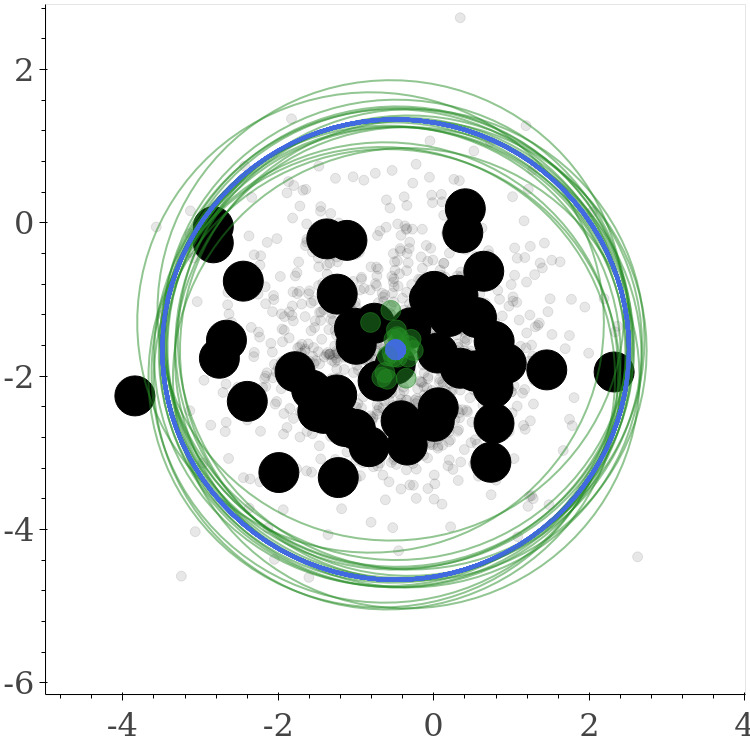}
\includegraphics[width=.32\columnwidth, clip, trim=100 100 80 80]{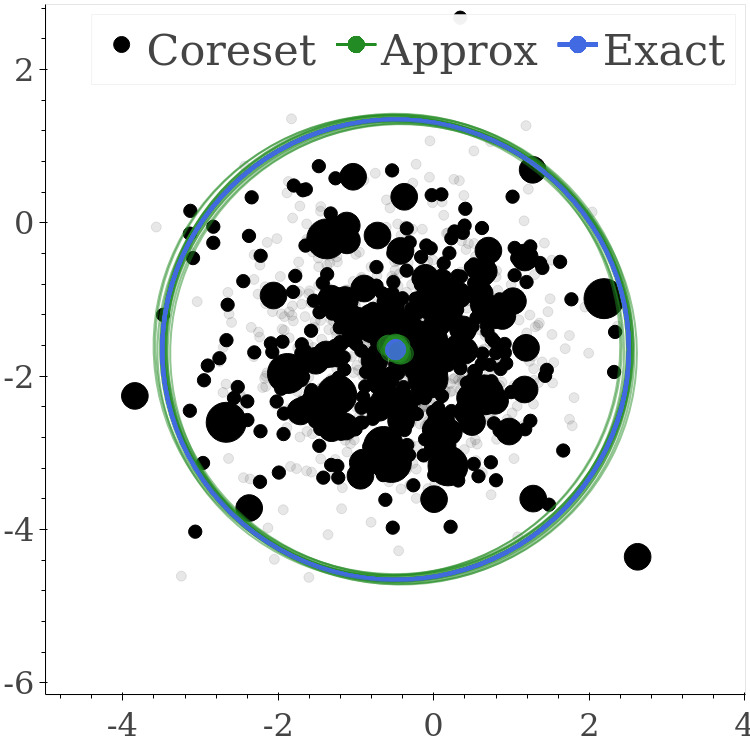}
\caption{Uniformly random subsampling}\label{fig:gauss_intuition_rand}
\end{subfigure}
\begin{subfigure}[t]{0.45\textwidth}
\includegraphics[width=.32\columnwidth, clip, trim=100 100 80 80]{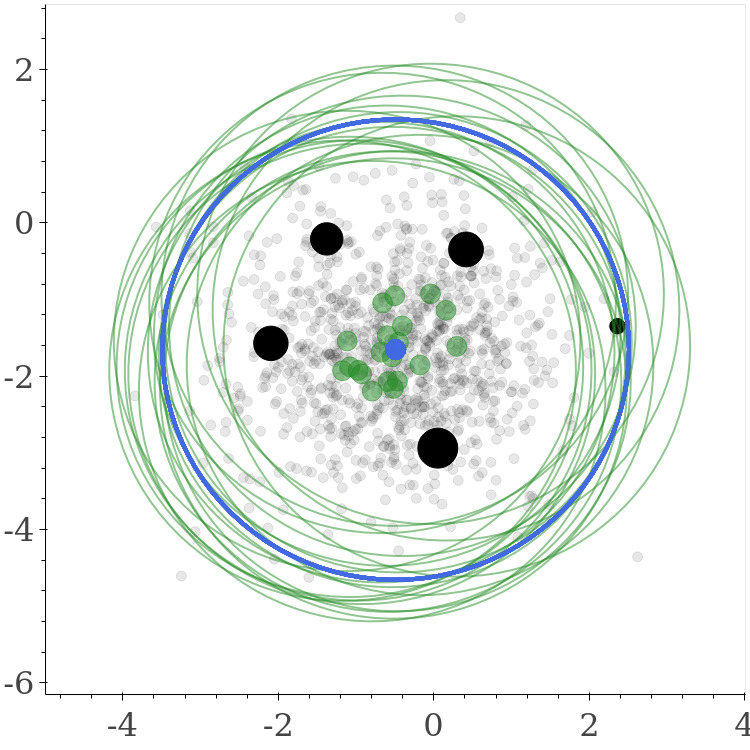}
\includegraphics[width=.32\columnwidth, clip, trim=100 100 80 80]{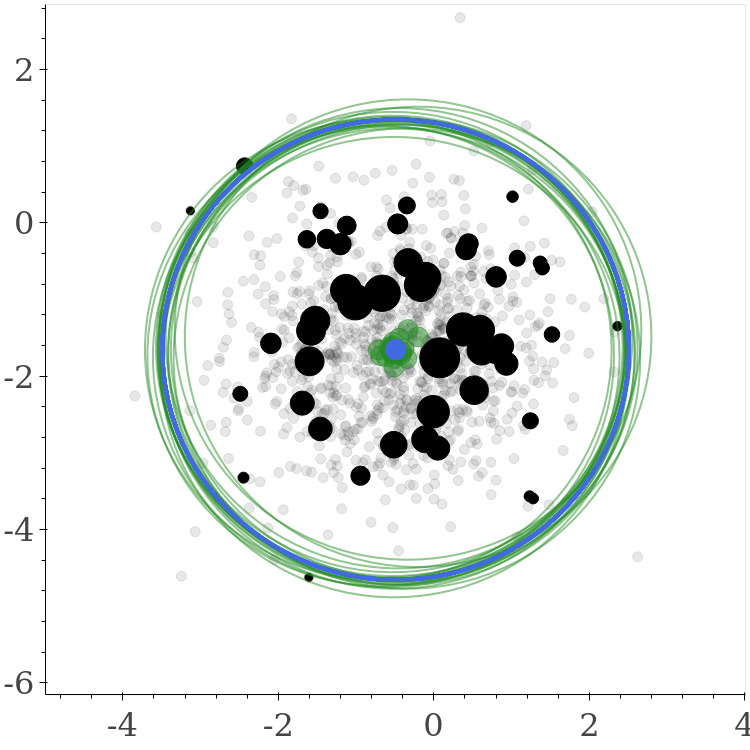}
\includegraphics[width=.32\columnwidth, clip, trim=100 100 80 80]{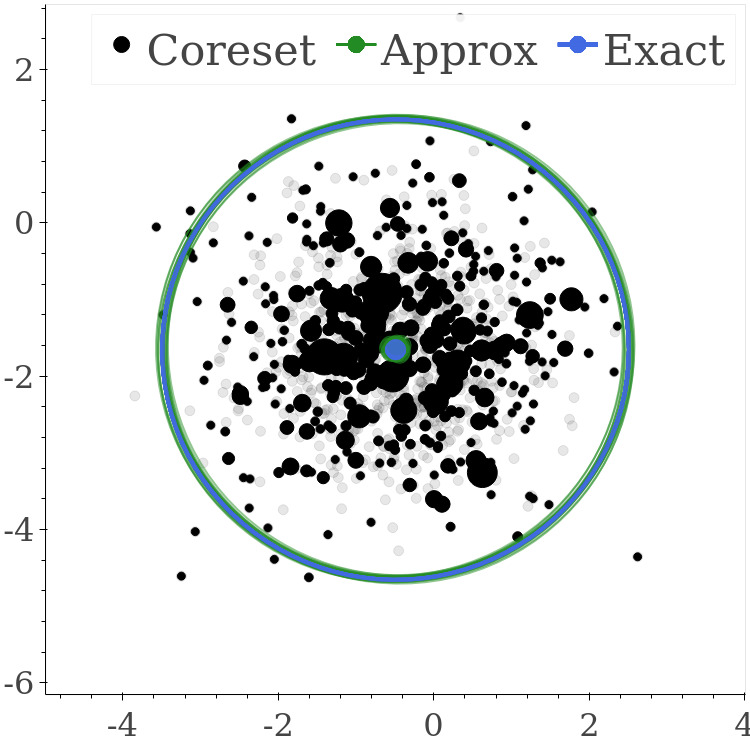}
\caption{Uniform coreset}\label{fig:gauss_intuition_unif}
\end{subfigure}
\begin{subfigure}[t]{0.45\textwidth}
\includegraphics[width=.32\columnwidth, clip, trim=100 100 80 80]{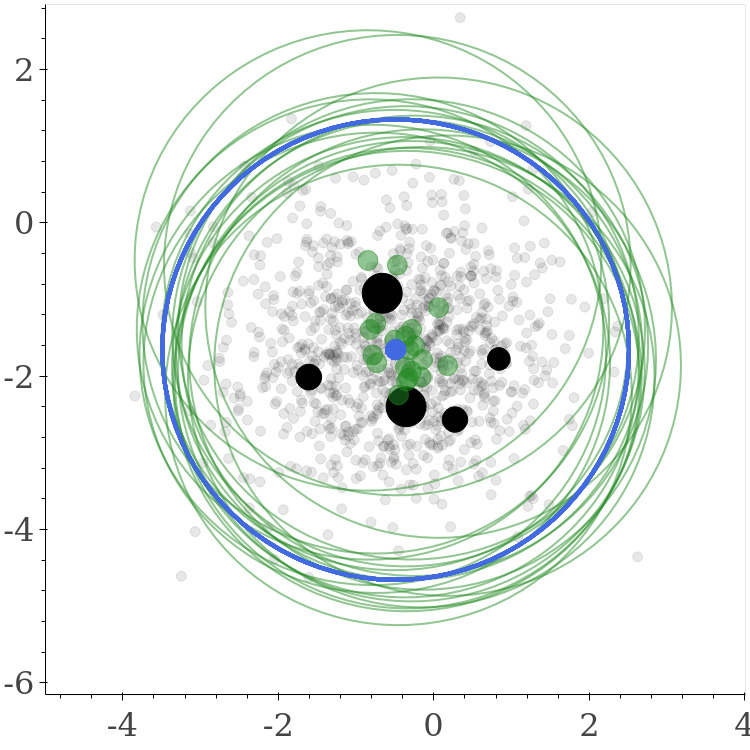}
\includegraphics[width=.32\columnwidth, clip, trim=100 100 80 80]{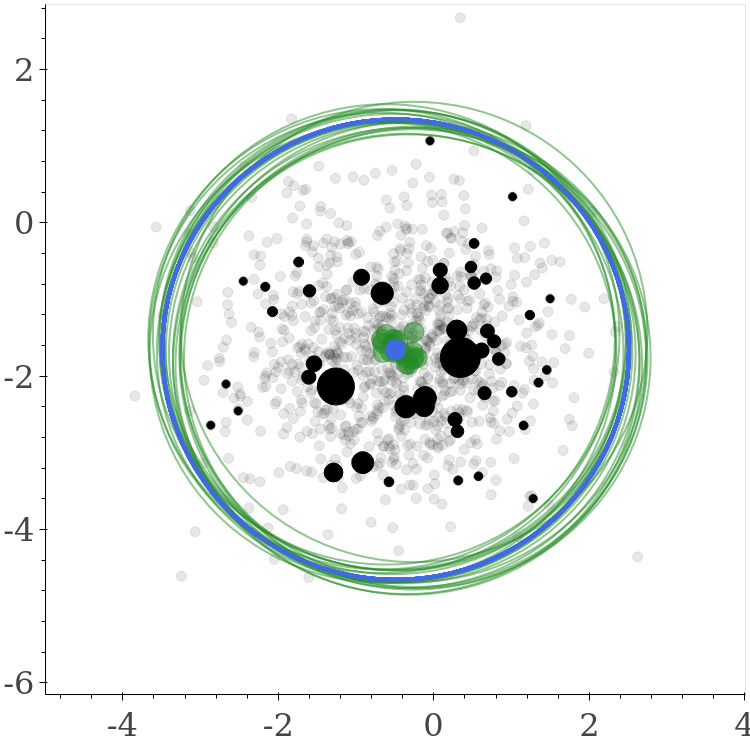}
\includegraphics[width=.32\columnwidth, clip, trim=100 100 80 80]{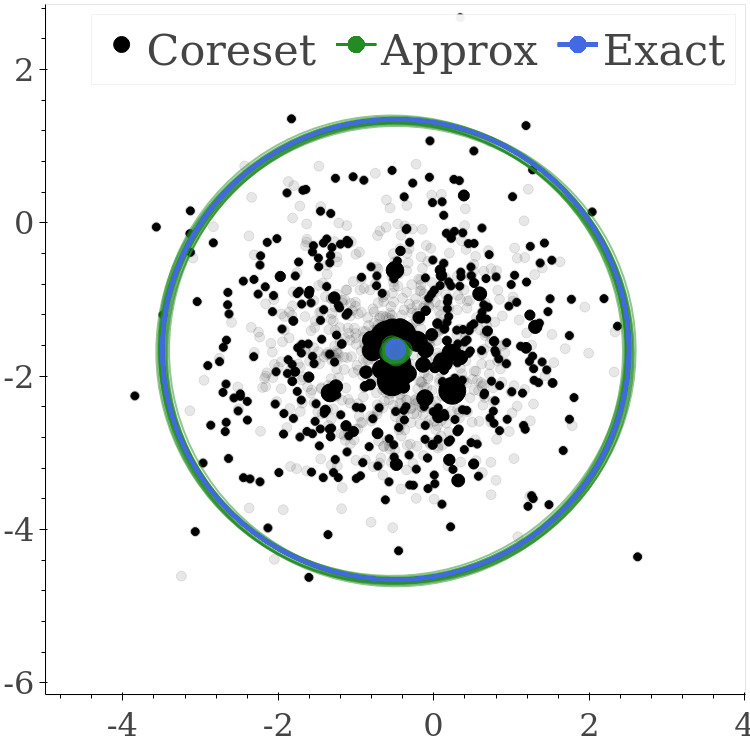}
\caption{Hilbert importance sampling}\label{fig:gauss_intuition_is}
\end{subfigure}
\begin{subfigure}[t]{0.45\textwidth}
\includegraphics[width=.32\columnwidth, clip, trim=100 100 80 80]{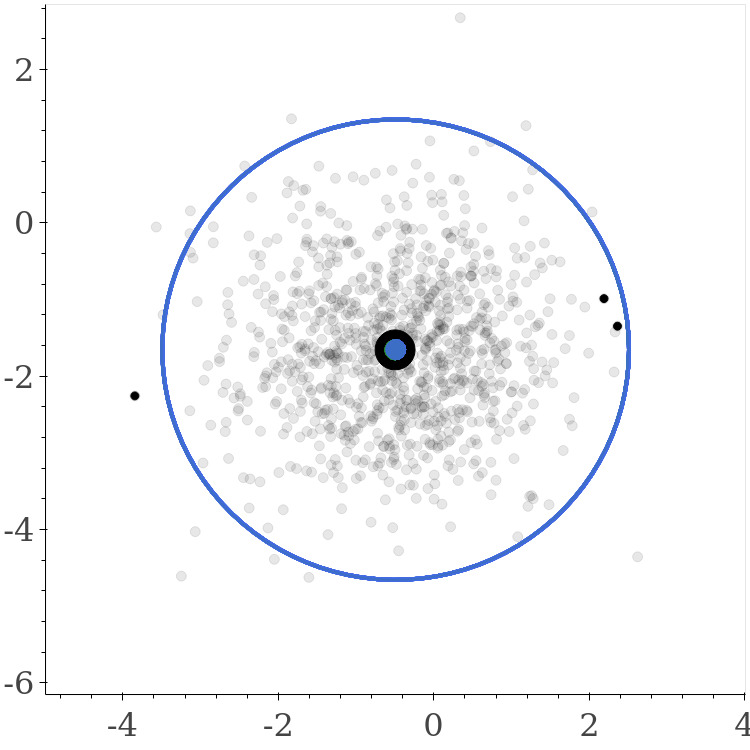}
\includegraphics[width=.32\columnwidth, clip, trim=100 100 80 80]{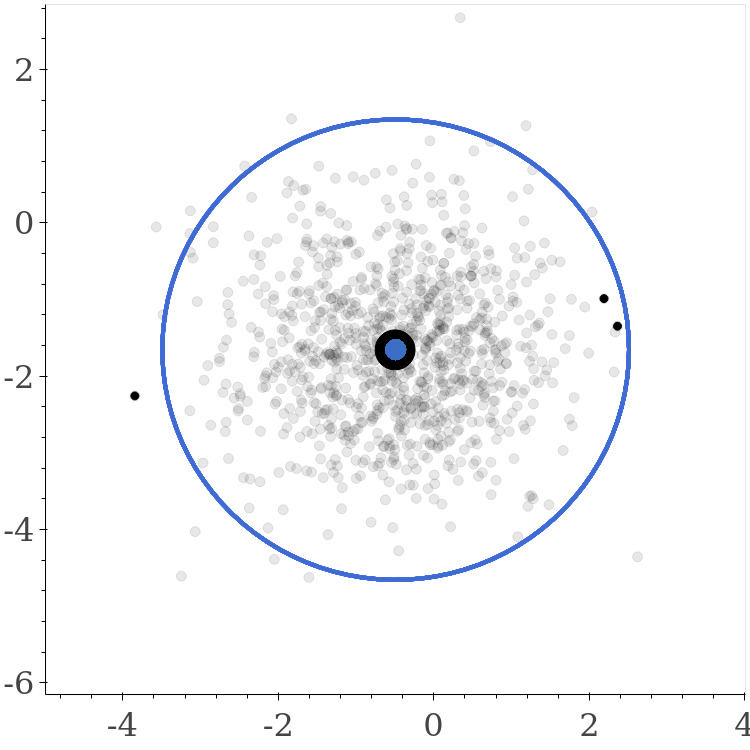}
\includegraphics[width=.32\columnwidth, clip, trim=100 100 80 80]{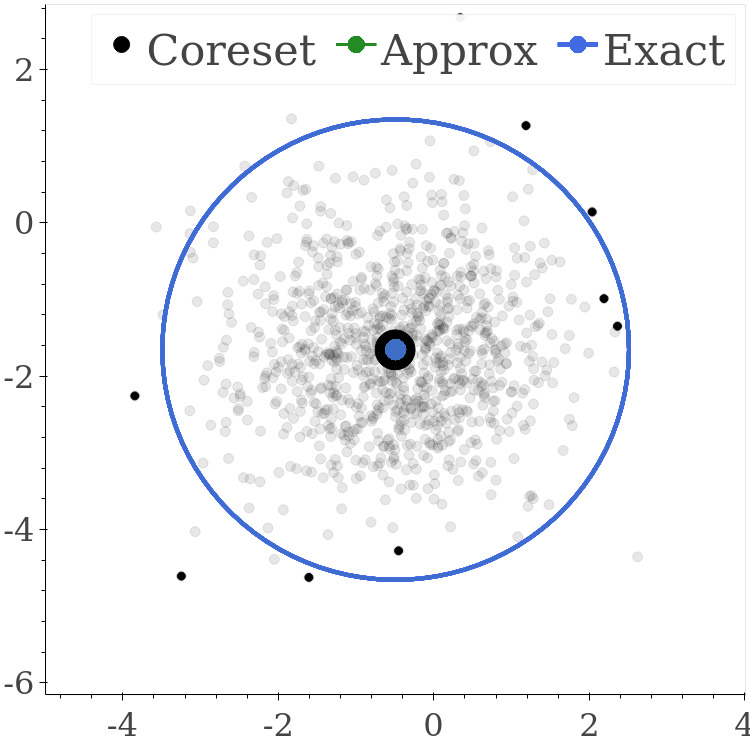}
\caption{Hilbert Frank--Wolfe}\label{fig:gauss_intuition_fw}
\end{subfigure}
\begin{subfigure}[t]{0.45\textwidth}
\includegraphics[width=\columnwidth]{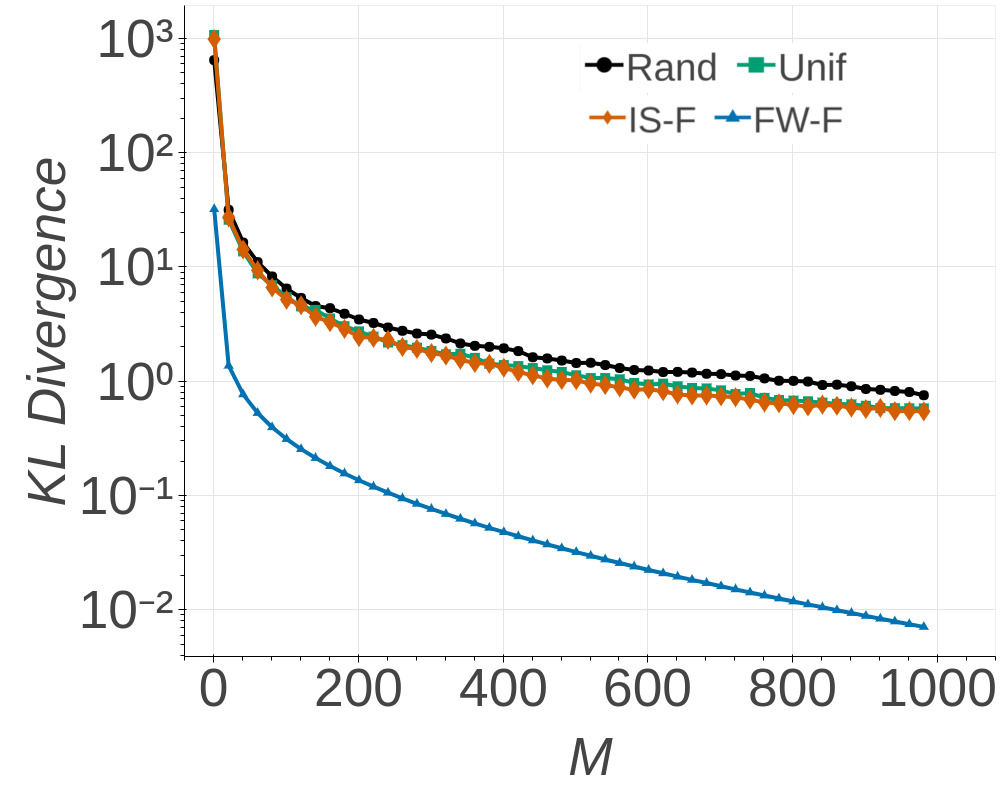}
\caption{}\label{fig:gauss_klM}
\end{subfigure}
\begin{subfigure}[t]{0.45\textwidth}
\includegraphics[width=\columnwidth]{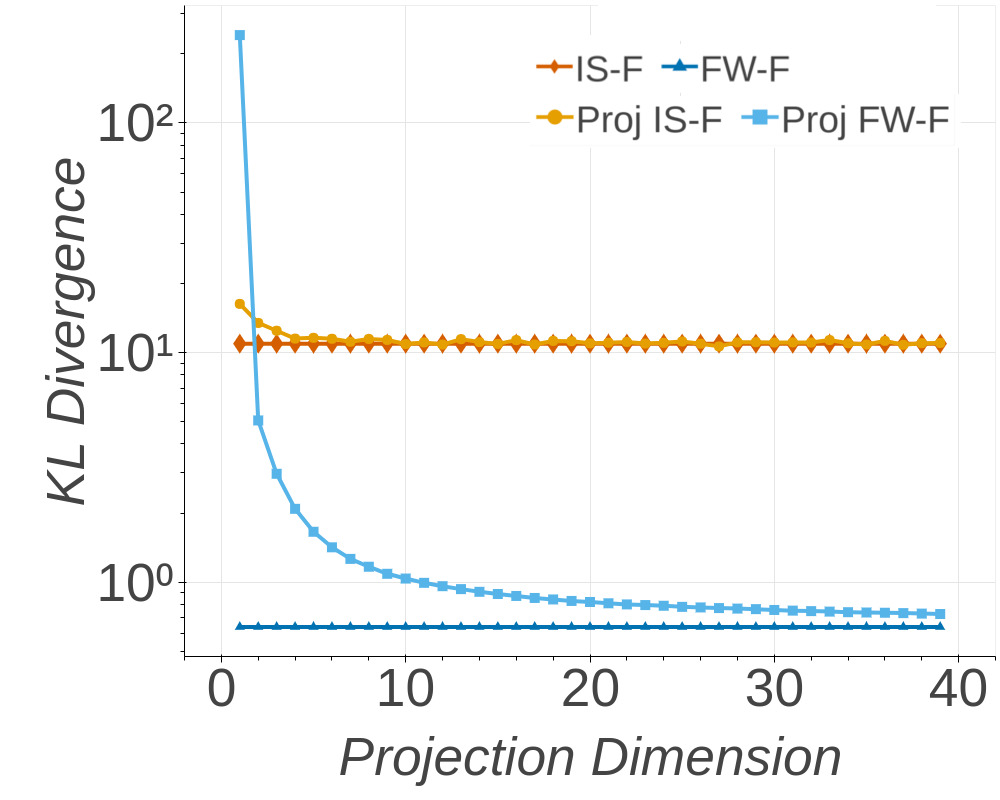}
\caption{}\label{fig:gauss_klD}
\end{subfigure}
\caption{(\ref{fig:gauss_intuition_rand}-\ref{fig:gauss_intuition_fw}):
Comparison of different coreset constructions for Gaussian inference, showing example coreset posterior 
predictive $3\sigma$ ellipses (green), the true data generating distribution $3\sigma$ ellipse (blue),
and a single trace of coreset construction (black) for $M =$ 5, 50, and 500. The radius of each 
coreset point indicates its weight.
(\ref{fig:gauss_klM}): A comparison of approximate posteriors using exact norms versus coreset construction iterations $M$.
(\ref{fig:gauss_klD}): A comparison of exact and projected methods versus projection dimension $J$, with fixed $M = 50$.
}
\label{fig:gauss_expt}
\end{figure}

\subsection*{Results}
The results of this test appear in \cref{fig:gauss_expt}.
The visual comparison of the different coreset constructions
in \cref{fig:gauss_intuition_rand}--\ref{fig:gauss_intuition_fw} makes the advantages of Hilbert coresets constructed via Frank--Wolfe clear.
As more coreset points are added, all approximate posteriors converge to the true posterior;
however, the Frank--Wolfe method requires many fewer coreset points to converge on a reliable estimate. While both the Hilbert and uniform coresets subsample the data favoring those points
at greater distance from the center, the Frank--Wolfe method first selects a point close to the center (whose scaled likelihood well-approximates the true posterior), and then refines
its estimate with points far away from the center. This intuitive behavior results in a higher-quality approximation of the posterior across all coreset sizes.
Note that the black coreset points across $M =$ 5, 50, and 500 show a single trace of a coreset being constructed, while the green posterior predictive ellipses show the noise
in the coreset construction across multiple runs at each fixed value of $M$.
The quantitative results in \cref{fig:gauss_klM,fig:gauss_klD}---which 
plot the KL-divergence between each coreset approximate posterior and the truth as the projection dimension $J$ or the number of coreset iterations $M$ 
varies---confirm the qualitative evaluations. In addition, \cref{fig:gauss_klM} confirms the theoretical result from \cref{thm:frankwolfe}, 
i.e., that Frank--Wolfe exhibits linear convergence in this setting.
\cref{fig:gauss_klD} similarly confirms \cref{thm:bayesfw}, i.e., that 
the posterior error of the projected Hilbert coreset converges to that of the exact Hilbert coreset
as the dimension $J$ of the random projection increases.

\section{Experiments}\label{sec:experiments}
In this section we evaluate the performance of Hilbert coresets compared with uniform coresets
and uniformly random subsampling, using MCMC on the full dataset as a benchmark. We test 
the algorithms on logistic regression, Poisson regression, and directional clustering models applied to numerous
real and synthetic datasets.
Based on the results of the synthetic comparison presented in \cref{sec:comparison}, for clarity we focus the tests
on comparing uniformly random subsampling to Hilbert coresets constructed using Frank--Wolfe with the weighted Fisher information distance from \cref{eq:normF}.
Additional results on importance sampling, uniform coresets, and the weighted 2-norm from \cref{eq:norm2}
are deferred to \cref{sec:additional_results}.

\subsection{Models}
In the \textbf{logistic regression} setting, we are given a set of data points $\left(x_n, y_n\right)_{n=1}^N$ each consisting of a feature $x_n\in\reals^D$
and a label $y_n\in\{-1, 1\}$, and the goal is to predict the label of a new point  given its feature.
We thus seek to infer the posterior distribution of the parameter $\theta\in\reals^{D+1}$ governing the generation of $y_n$ given $x_n$
via
\[
\theta &\dist\distNorm(0, I) & y_n \given x_n, \theta &\distind \distBern\left(\frac{1}{1+e^{-z_n^T\theta}}\right) & z_n &\defined \left[x_n, \, 1\right]^T.
\]
In the \textbf{Poisson regression} setting, we are given a set of data points $\left(x_n, y_n\right)_{n=1}^N$ each consisting of a feature $x_n\in\reals^D$
and a count $y_n\in\nats$, and the goal is to learn a relationship between features $x_n$ and the associated mean count.
We thus seek to infer the posterior distribution of the parameter $\theta\in\reals^D$ governing the generation of $y_n$ given $x_n$
via
\[
\theta &\dist\distNorm(0, I) & y_n \given x_n, \theta &\distind \distPoiss\left(\log\left(1+e^{\theta^Tz_n}\right)\right) & z_n &\defined \left[x_n, 1\right]^T.
\]
Finally, in the \textbf{directional clustering} setting, we are given a dataset of points $\left(x_n\right)_{n=1}^N$ on the unit $(D-1)$-sphere, i.e.~$x_n \in \reals^D$ with $\|x_n\|_2 = 1$,
and the goal is to separate them into $K$ clusters. For this purpose we employ a \emph{von Mises-Fisher (vMF) mixture model} \citep{Banerjee05}.
The component likelihood in this model is the von Mises-Fisher distribution $\distVMF(\mu, \tau)$ with concentration $\tau \in \reals_+$
and mode $\mu \in \reals^D$, $\|\mu\|_2 = 1$,  having density
\[
f_{\distVMF}(x; \mu, \tau) &= C_D(\tau)e^{\tau x^T\mu} & C_D(\tau) &= \frac{\tau^{D/2-1}}{(2\pi)^{D/2}I_{D/2-1}(\tau)} \label{eq:vmfdensity}
\]
with support on the unit $(D-1)$-sphere $\mathbb{S}^{D-1}$,
where $I_p$ denotes the modified Bessel function of the first kind of order $p$. We place uniform priors on both the component modes
and mixture weights, and set $\tau = 50$, resulting in the generative model
\[
&(\mu_k)_{k=1}^K \distiid \distUnif\left(\mathbb{S}^{D-1}\right) \quad (\omega_k)_{k=1}^K \dist \distDir(1, \dots, 1)\\
  &(x_n)_{n=1}^N\given (\omega_k, \mu_k)_{k=1}^K \distiid \sum_{k=1}^K \omega_k \distVMF(\mu_{k}, \tau).  
\]

\subsection{Datasets}
\begin{figure}[t!]
\begin{subfigure}[t]{0.45\textwidth}
\includegraphics[width=.32\columnwidth, clip, trim=30 30 32 32]{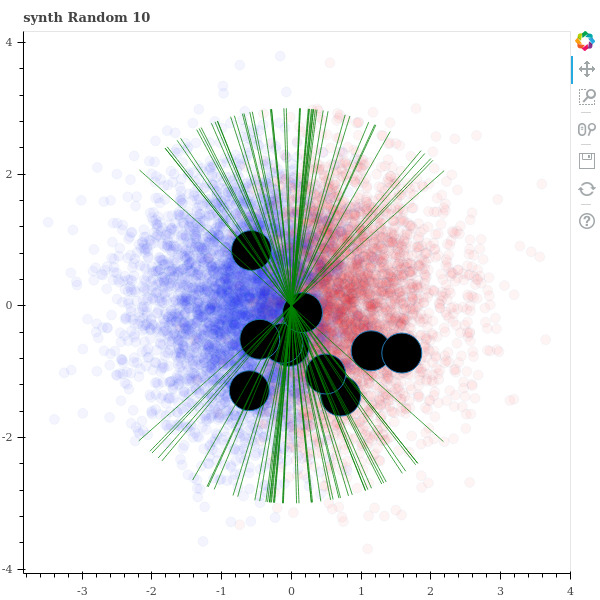}
\includegraphics[width=.32\columnwidth, clip, trim=30 30 32 32]{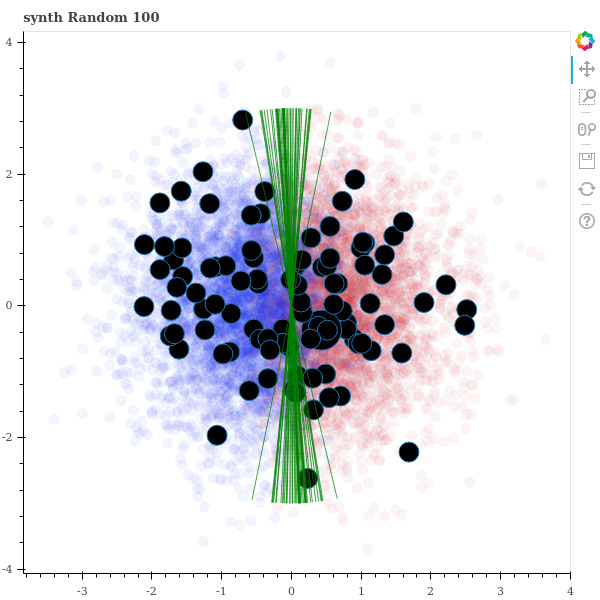}
\includegraphics[width=.32\columnwidth, clip, trim=30 30 32 32]{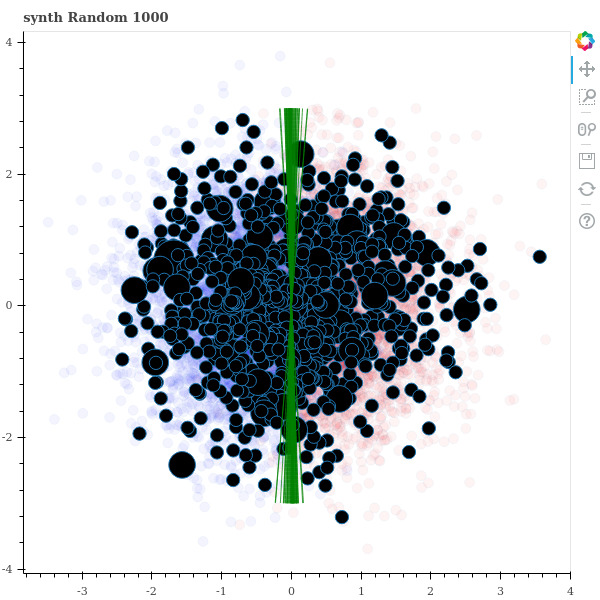}
\caption{Uniformly random subsampling}\label{fig:lr_intuition_rand}
\end{subfigure}
\begin{subfigure}[t]{0.45\textwidth}
\includegraphics[width=.32\columnwidth, clip, trim=30 30 32 32]{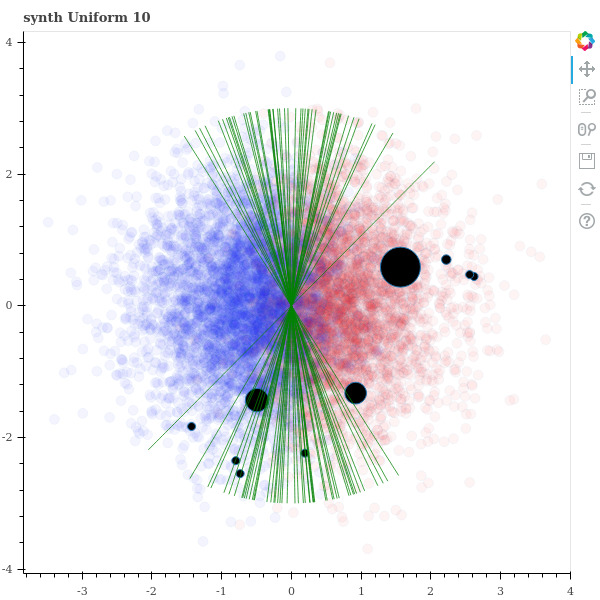}
\includegraphics[width=.32\columnwidth, clip, trim=30 30 32 32]{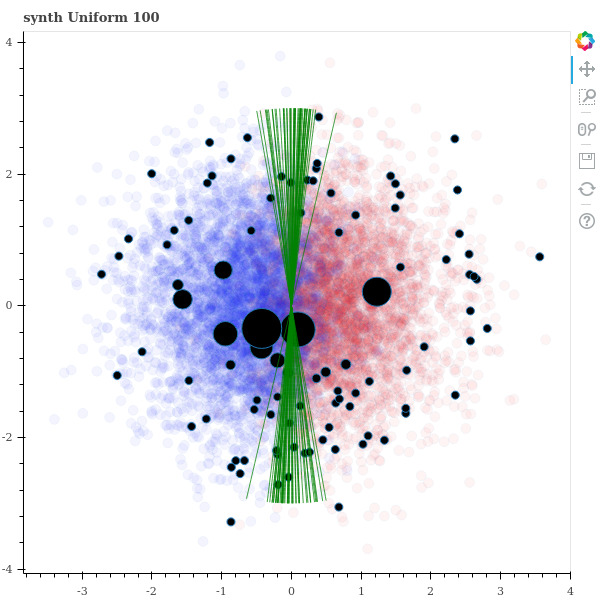}
\includegraphics[width=.32\columnwidth, clip, trim=30 30 32 32]{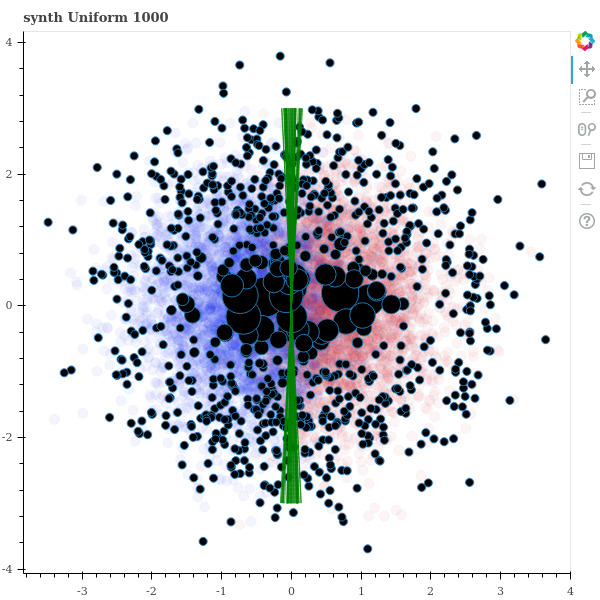}
\caption{Uniform coreset}
\end{subfigure}
\begin{subfigure}[t]{0.45\textwidth}
\includegraphics[width=.32\columnwidth, clip, trim=30 30 32 32]{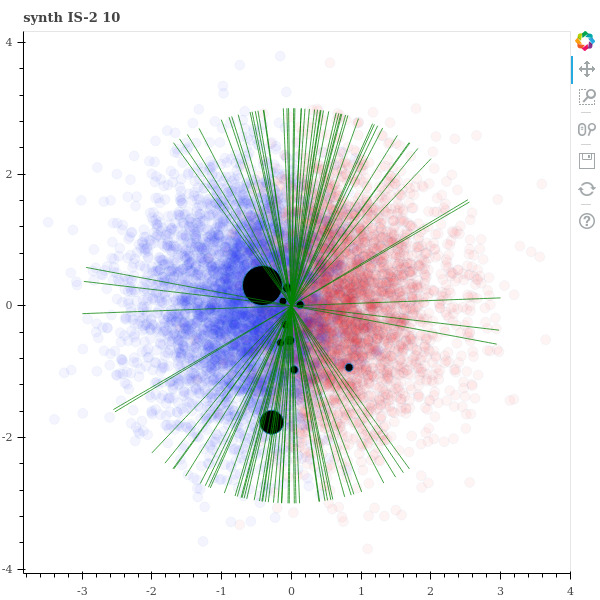}
\includegraphics[width=.32\columnwidth, clip, trim=30 30 32 32]{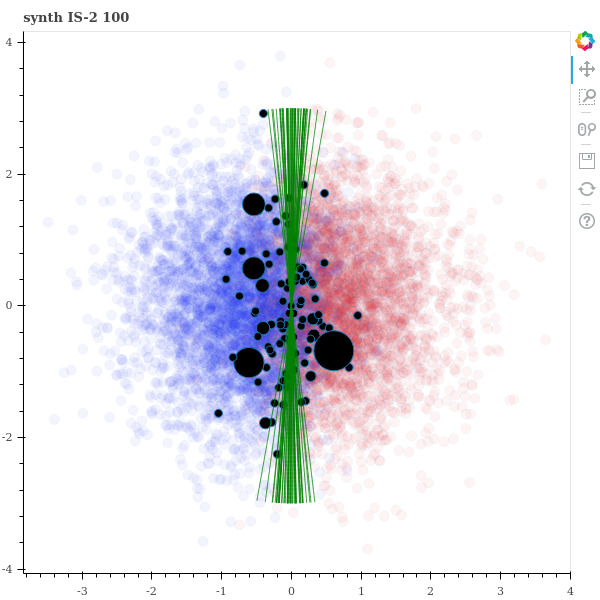}
\includegraphics[width=.32\columnwidth, clip, trim=30 30 32 32]{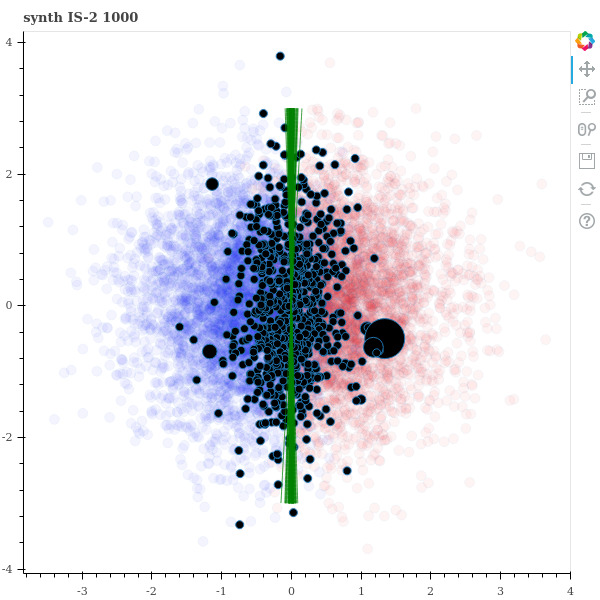}
\caption{Hilbert importance sampling}
\end{subfigure}
\begin{subfigure}[t]{0.45\textwidth}
\includegraphics[width=.32\columnwidth, clip, trim=30 30 32 32]{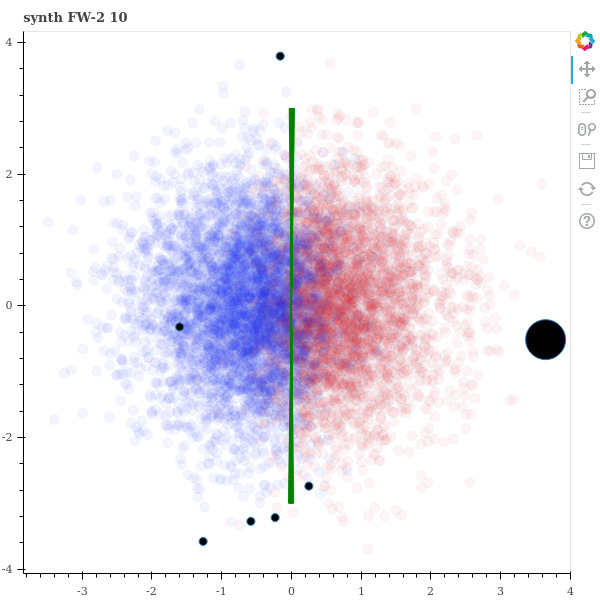}
\includegraphics[width=.32\columnwidth, clip, trim=30 30 32 32]{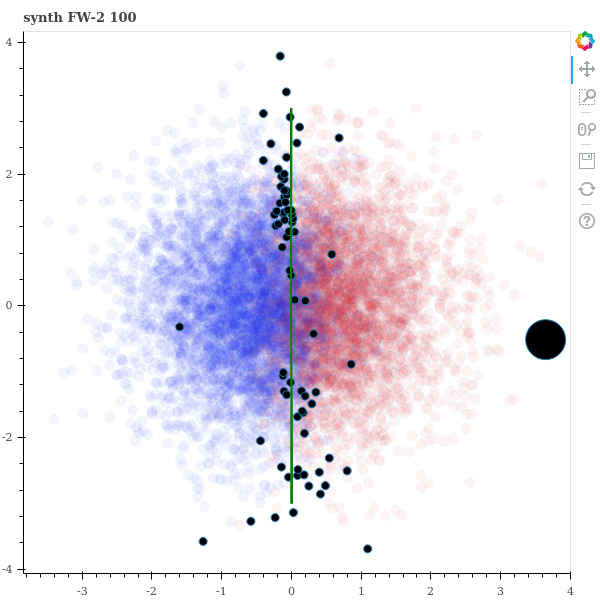}
\includegraphics[width=.32\columnwidth, clip, trim=30 30 32 32]{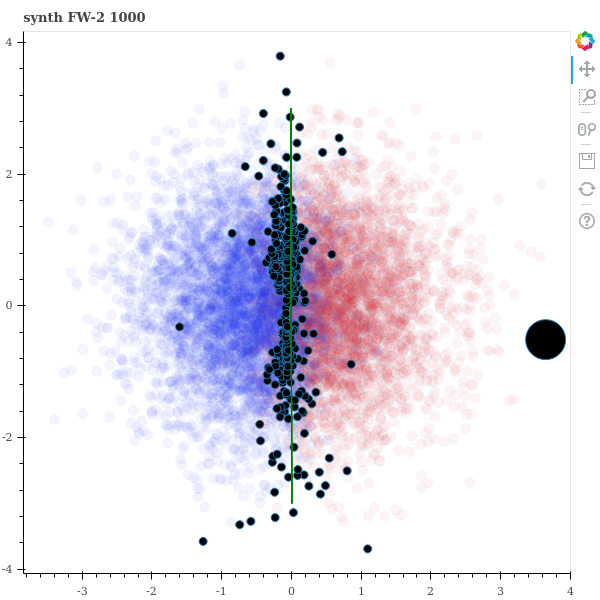}
\caption{Hilbert Frank--Wolfe}\label{fig:lr_intuition_fw}
\end{subfigure}
\begin{subfigure}[t]{0.45\textwidth}
\includegraphics[width=\columnwidth]{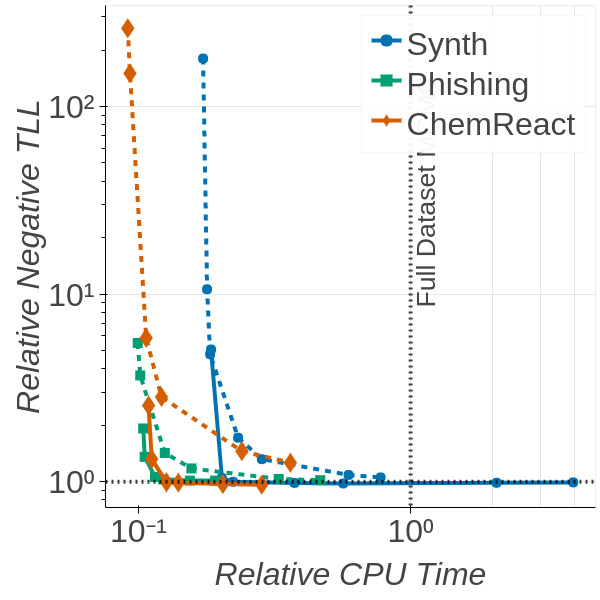}
\caption{}\label{fig:lr_tll}
\end{subfigure}
\begin{subfigure}[t]{0.45\textwidth}
\includegraphics[width=\columnwidth]{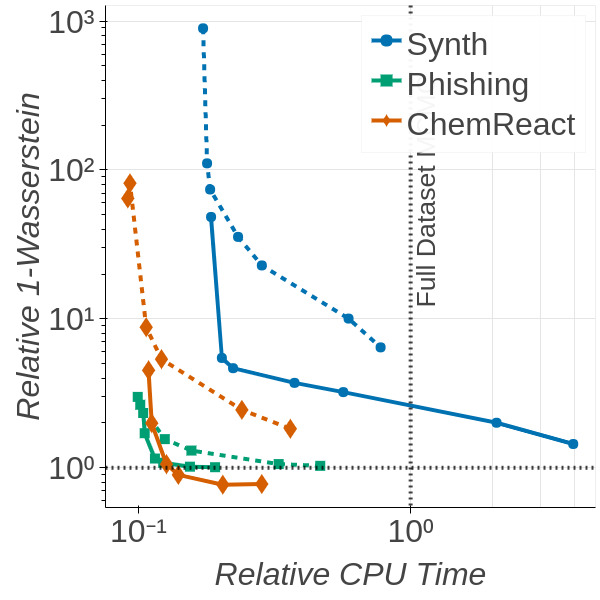}
\caption{}\label{fig:lr_w1}
\end{subfigure}
\caption{(\ref{fig:lr_intuition_rand}-\ref{fig:lr_intuition_fw}): Comparison of different coreset constructions for logistic regression on the \texttt{Synthetic} dataset (with blue \& red labeled data), 
showing example coreset posterior mean classification boundaries (green),
and a single trace of coreset construction (black) for $M =$ 10, 100, and 1000. The radius of each coreset point indicates its weight.
(\ref{fig:lr_tll}, \ref{fig:lr_w1}): A comparison of negative test log-likelihood (\ref{fig:lr_tll}) and 1-Wasserstein distance (\ref{fig:lr_w1})
versus computation time for Frank--Wolfe (solid) and uniform random subsampling (dashed) on the logistic regression model.
Both axes are normalized using results from running MCMC on the full dataset; see \cref{sec:expt_methods}.
}\label{fig:lr_expt}
\end{figure}

We tested the coreset construction methods for each model on a number of datasets.
For \textbf{logistic regression}, the \texttt{Synthetic} dataset consisted of $N=$ 10,000 data points (with 1,000 held out for testing) with covariate $x_n \in \reals^2$ 
sampled \iid from $\distNorm(0, I)$, and label $y_n\in\{-1, 1\}$ generated from the logistic likelihood with parameter $\theta = \left[3, 3, 0\right]^T$.  
The \texttt{Phishing}\footnote{\url{https://www.csie.ntu.edu.tw/~cjlin/libsvmtools/datasets/binary.html}} dataset
consisted of $N=$ 11,055 data points (with 1,105 held out for testing) each with $D=$ 68 features. 
In this dataset, each covariate corresponds to the features of a website, and the goal is to predict whether or not a website is a phishing site.
The \texttt{ChemReact}\footnote{\url{http://komarix.org/ac/ds/}} dataset 
consisted of $N=$ 26,733 data points (with 2,673 held out for testing) each with $D=$ 10 features. In this dataset, each covariate represents the features of a 
chemical experiment, and the label represents whether a chemical was reactive in that experiment or not. 

For \textbf{Poisson regression}, the \texttt{Synthetic} dataset consisted of $N=$ 10,000 data points (with 1,000 held out for testing) with covariate $x_n \in \reals$ 
sampled \iid from $\distNorm(0, 1)$, and count $y_n\in\nats$ generated from the Poisson likelihood with $\theta = \left[1, 0\right]^T$.
The \texttt{BikeTrips}\footnote{\url{http://archive.ics.uci.edu/ml/datasets/Bike+Sharing+Dataset}} dataset
consisted of $N=$ 17,386 data points (with 1,738 held out for testing) each with $D=$ 8 features. 
In this dataset, each covariate corresponds to weather and season information for a particular hour during the time between 2011--2012, and the count is the number of bike trips
taken during that hour in a bikeshare system in Washington, DC.
The \texttt{AirportDelays}\footnote{Airport information from \url{http://stat-computing.org/dataexpo/2009/the-data.html}, with historical weather information from \url{https://www.wunderground.com/history/}.} dataset 
consisted of $N=$ 7,580 data points (with 758 held out for testing) each with $D=$ 15 features. In this dataset, each covariate 
corresponds to the weather information of a day during the time between 1987--2008, and
the count is the number of flights leaving Boston Logan airport delayed by more than 15 minutes that day. 

Finally, for \textbf{directional clustering}, the \texttt{Synthetic} dataset consisted of $N=$ 10,000 data points (with 1,000 held out for testing) generated from an equally-weighted $\distVMF$ mixture
with 6 components, one centered at each of the axis poles.

\subsection{Methods}\label{sec:expt_methods}
\begin{figure}[t!]
\begin{center}
\begin{subfigure}[t]{0.45\textwidth}
\includegraphics[width=.32\columnwidth, clip, trim=40 40 40 40]{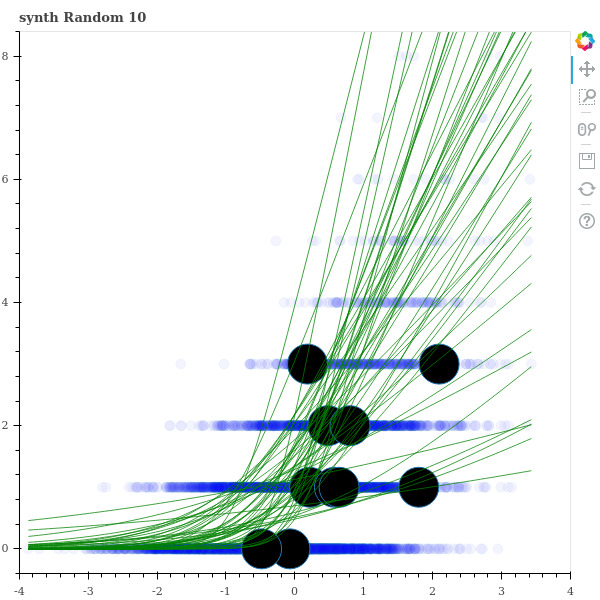}
\includegraphics[width=.32\columnwidth, clip, trim=40 40 40 40]{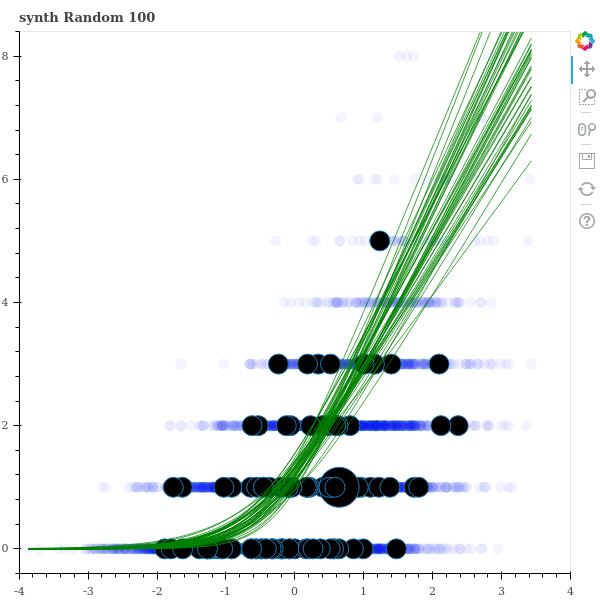}
\includegraphics[width=.32\columnwidth, clip, trim=40 40 40 40]{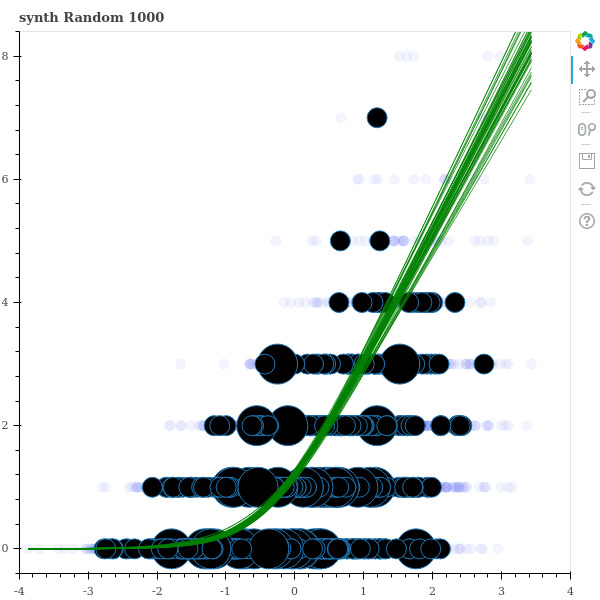}
\caption{Uniformly random subsampling}\label{fig:poiss_intuition_rand}
\end{subfigure}
\end{center}
\begin{subfigure}[t]{0.45\textwidth}
\includegraphics[width=.32\columnwidth, clip, trim=40 40 40 40]{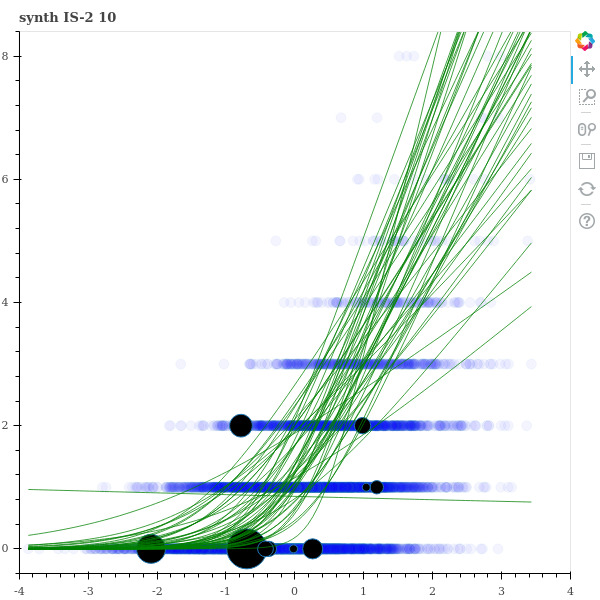}
\includegraphics[width=.32\columnwidth, clip, trim=40 40 40 40]{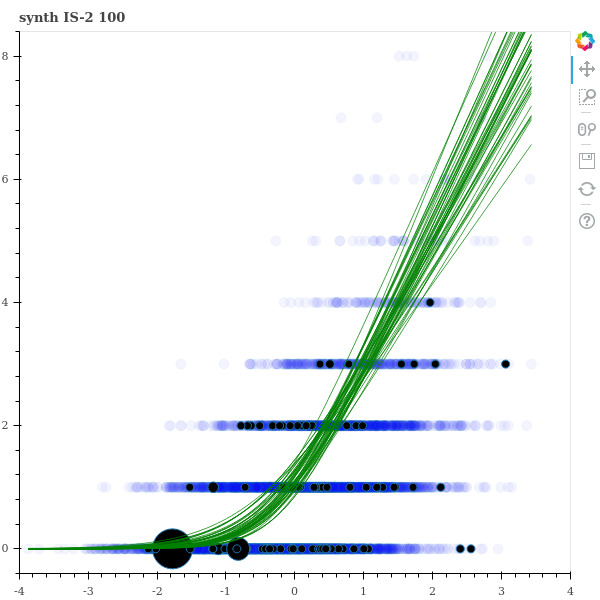}
\includegraphics[width=.32\columnwidth, clip, trim=40 40 40 40]{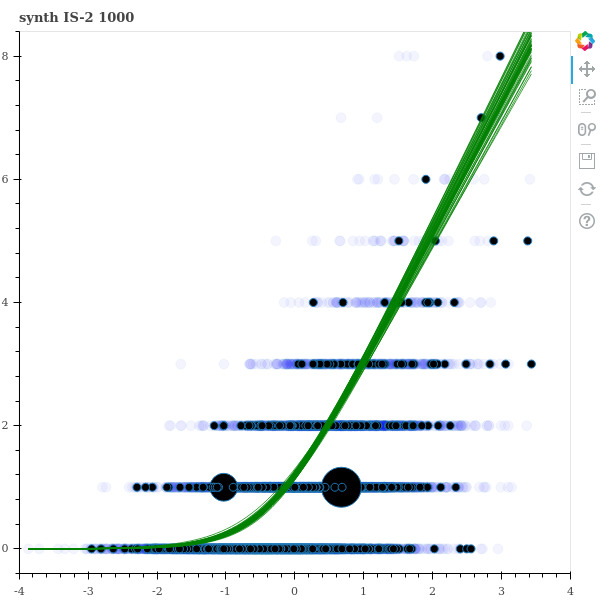}
\caption{Hilbert importance sampling}
\end{subfigure}
\begin{subfigure}[t]{0.45\textwidth}
\includegraphics[width=.32\columnwidth, clip, trim=40 40 40 40]{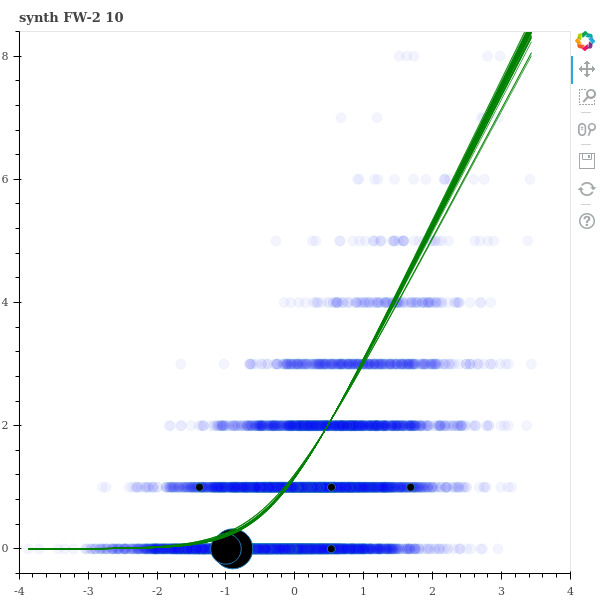}
\includegraphics[width=.32\columnwidth, clip, trim=40 40 40 40]{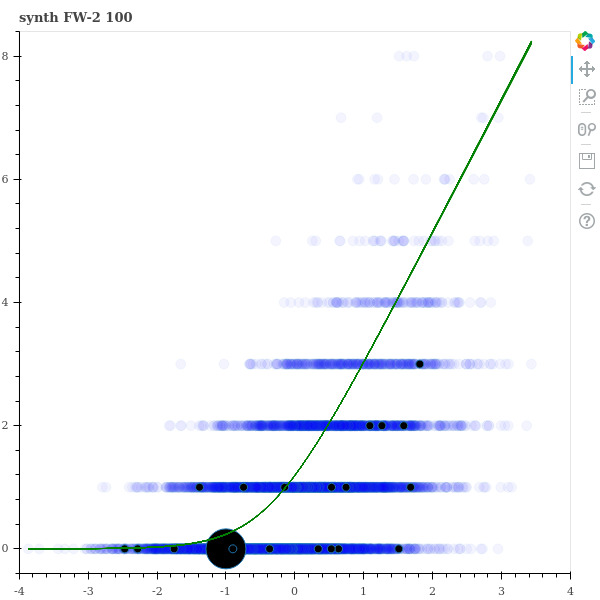}
\includegraphics[width=.32\columnwidth, clip, trim=40 40 40 40]{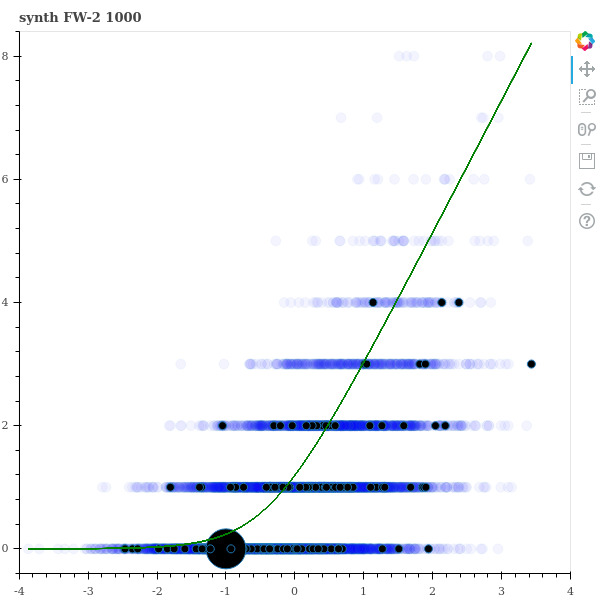}
\caption{Hilbert Frank--Wolfe}\label{fig:poiss_intuition_fw}
\end{subfigure}
\begin{subfigure}[t]{0.45\textwidth}
\includegraphics[width=\columnwidth]{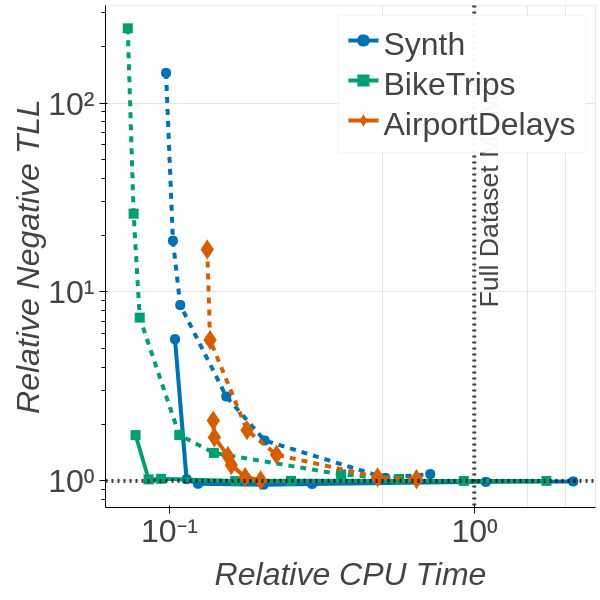}
\caption{}\label{fig:poiss_tll}
\end{subfigure}
\begin{subfigure}[t]{0.45\textwidth}
\includegraphics[width=\columnwidth]{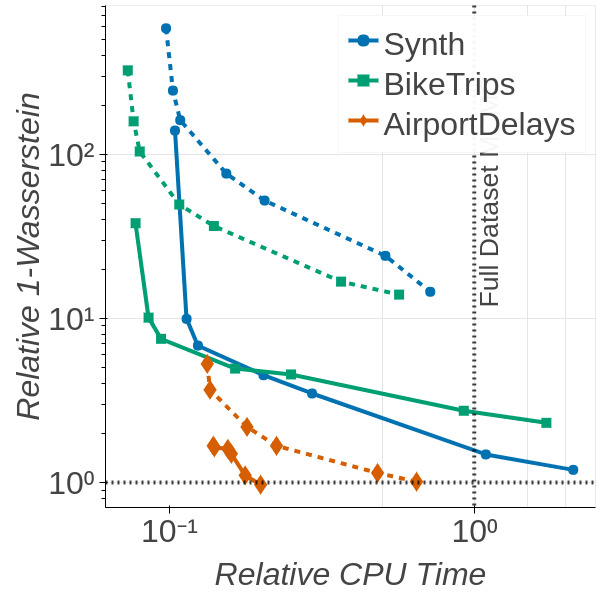}
\caption{}\label{fig:poiss_w1}
\end{subfigure}
\caption{(\ref{fig:poiss_intuition_rand}-\ref{fig:poiss_intuition_fw}): Comparison of different coreset constructions for Poisson regression on the \texttt{Synthetic} dataset,
 showing example coreset posterior Poisson mean curves (green),
and a single trace of coreset construction (black) for $M =$ 10, 100, and 1,000. The radius of each coreset point indicates its weight.
(\ref{fig:poiss_tll}, \ref{fig:poiss_w1}): A comparison of negative test log-likelihood (\ref{fig:poiss_tll}) and 1-Wasserstein distance (\ref{fig:poiss_w1})
versus computation time for Frank--Wolfe (solid) and uniform random subsampling (dashed) on the Poisson regression model.
Both axes are normalized using results from running MCMC on the full dataset; see \cref{sec:expt_methods}.
}\label{fig:poiss_expt}
\end{figure}

We ran 50 trials of uniformly random subsampling and 
Hilbert Frank--Wolfe using the approximate Fisher information distance in \cref{eq:normF},
varying $M \in \{$10, 50, 100, 500, 1,000, 5,000, 10,000$\}$.
For both \textbf{logistic regression} and \textbf{Poisson regression}, we used the Laplace approximation \citep[Section 4.4]{Bishop06} as the weighting distribution $\hat\pi$ 
in the Hilbert coreset, with the random projection dimension set to $D=$ 500. Posterior inference in each of the 50 trials 
was conducted using random walk Metropolis-Hastings with an isotropic multivariate Gaussian proposal distribution.
We simulated a total of 100,000 steps, with 50,000 warmup steps including proposal covariance adaptation with a target acceptance rate of 0.234, and thinning of the latter 50,000
by a factor of 5, yielding 10,000 posterior samples.

For \textbf{directional clustering} the weighting distribution $\hat\pi$ for the Hilbert coreset was constructed
by finding maximum likelihood estimates of the cluster modes $(\hat\mu_k)_{k=1}^K$ and weights $\hat\omega$
using the EM algorithm, and then setting $\hat\pi$ to an independent product of approximate posterior conditionals,
\[
\<
\bx_k &\defined \sum_{n=1}^{N}z_{nk}x_n & \bz_k &\defined \sum_{n=1}^N z_{nk}\\
\mu_k &\distind \distVMF\left(\frac{\bx_k}{\|\bx_k\|}, \tau\|\bx_k\|\right) &
\omega &\distind \distDir\left(1+\bz_1, \dots, 1+\bz_k\right),
\>\label{eq:vmfsamplewtsctrs}
\]
where $(z_n)_{n=1}^N$, $z_n\in\reals_+^K$ are the smoothed cluster assignments.
The random projection dimension was set to $D=$ 500, and the number of clusters $K$ was set to 6.
Posterior inference in each of the 50 trials was conducted used Gibbs sampling (introducing auxiliary label variables for the data) with a total of 100,000 steps, with 50,000 warmup steps and thinning of the latter 50,000
by a factor of 5, yielding 10,000 posterior samples. Note that this approach is exact for the full dataset; for the coreset constructions with weighted data, we replicate each data point by its ceiled weight,
and then rescale the assignment variables to account for the fractional weight. 
In particular, for coreset weights $(w_n)_{n=1}^N$, we sample labels for points with $w_n > 0$ via
\[
\gamma_k &\propto \omega_k f_{\distVMF}\left(x_n; \mu_k, \tau\right) \quad z_n \distind \distMulti(\texttt{ceil}(w_n), \gamma) \quad z_n \gets \frac{w_n}{\texttt{ceil}(w_n)}z_n,
\]
and sample the cluster centers and weights via \cref{eq:vmfsamplewtsctrs}.

For all models, we evaluate two metrics of posterior quality: negative log-likelihood on the held-out test set, averaged over posterior 
MCMC samples; and 1-Wasserstein distance of the posterior samples to samples obtained from running MCMC on the full dataset. 
All negative test log-likelihood results are shifted by the maximum possible test log-likelihood and 
normalized by the test log-likelihood obtained from the full dataset posterior. All 1-Wasserstein distance results
are normalized by the median pairwise 1-Wasserstein distance between 10 trials of MCMC on the full dataset.
All computation times are normalized by the median computation time for MCMC on the full dataset across the 10 trials. 
These normalizations allow the results from multiple datasets to be plotted coherently on the same axes.

We ran the same experiments described above on Hilbert importance sampling for all datasets,
and uniform coresets on the logistic regression model with $a=3$ and $K=4$ (see \citet[Sec.~4.2]{Huggins16}).
We also compared Hilbert coresets with the weighted 2-norm in \cref{eq:norm2} to the weighted Fisher information distance in \cref{eq:normF}.
The results of these experiments are deferred to \cref{sec:additional_results} for clarity.

\subsection{Results and discussion}\label{sec:expt_results}
\begin{figure}[t!]
\begin{center}
\begin{subfigure}[t]{0.45\textwidth}
\includegraphics[width=.32\columnwidth, clip, trim=40 40 40 40]{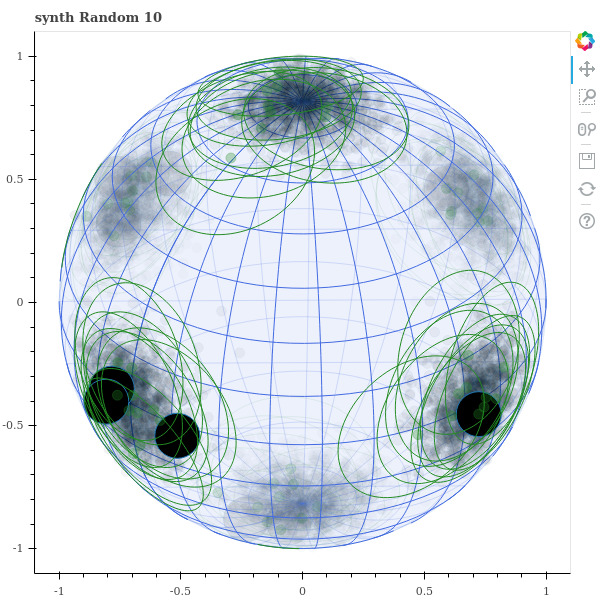}
\includegraphics[width=.32\columnwidth, clip, trim=40 40 40 40]{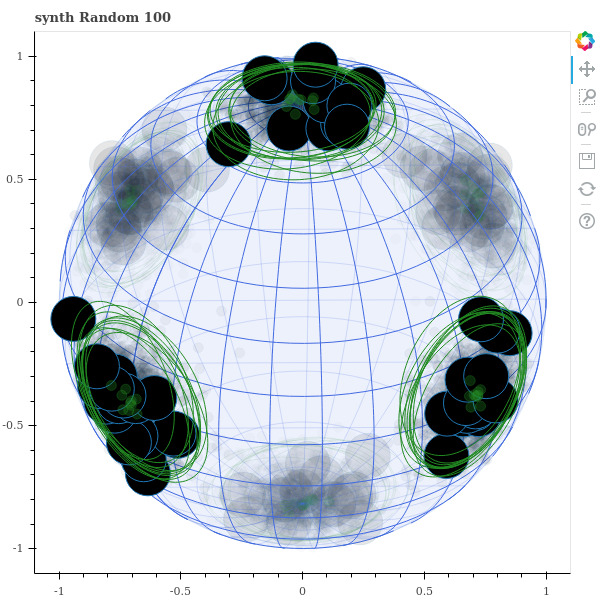}
\includegraphics[width=.32\columnwidth, clip, trim=40 40 40 40]{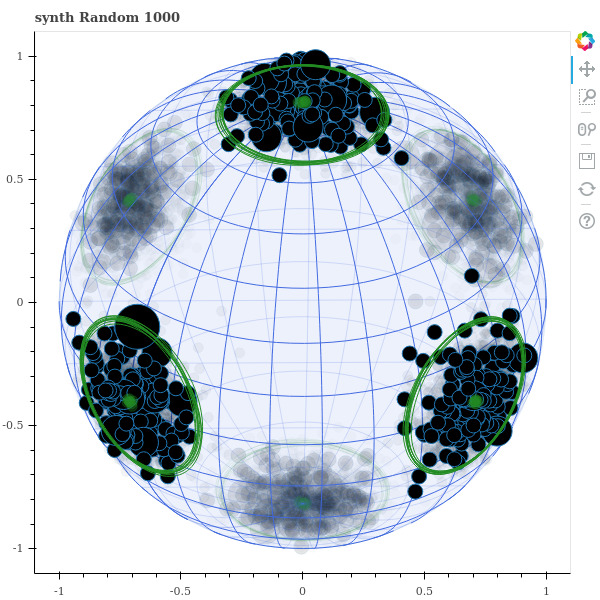}
\caption{Uniformly random subsampling}\label{fig:vmf_intuition_rand}
\end{subfigure}
\end{center}
\begin{subfigure}[t]{0.45\textwidth}
\includegraphics[width=.32\columnwidth, clip, trim=40 40 40 40]{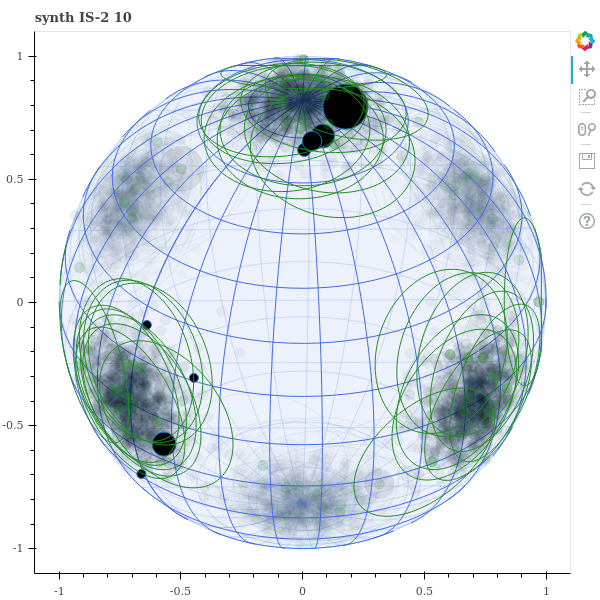}
\includegraphics[width=.32\columnwidth, clip, trim=40 40 40 40]{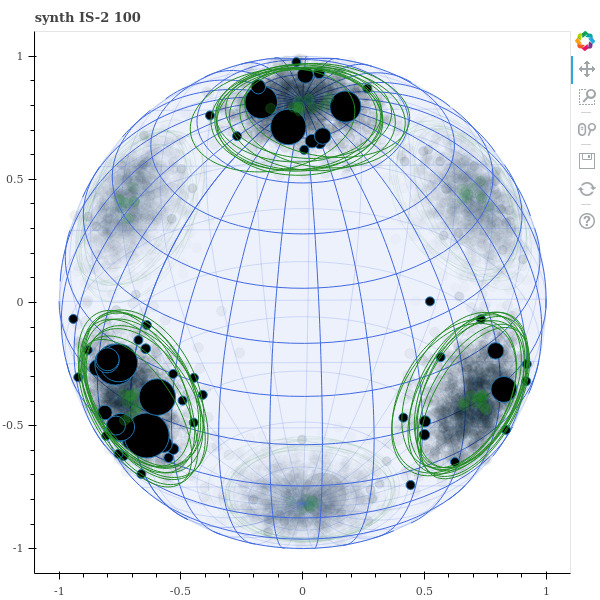}
\includegraphics[width=.32\columnwidth, clip, trim=40 40 40 40]{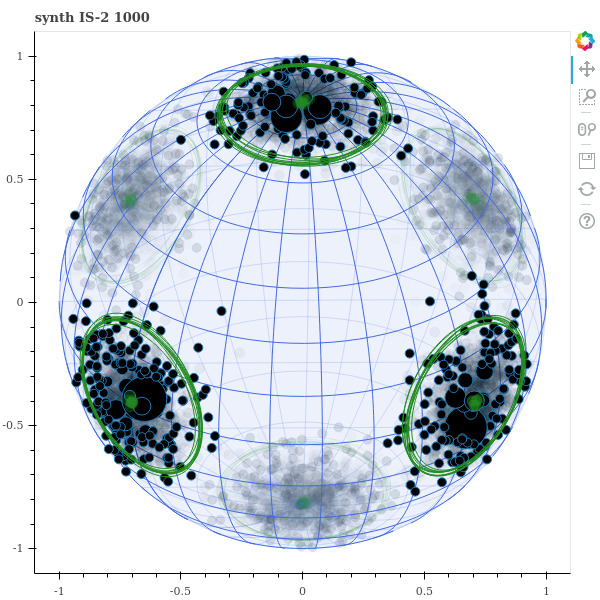}
\caption{Hilbert importance sampling}
\end{subfigure}
\begin{subfigure}[t]{0.45\textwidth}
\includegraphics[width=.32\columnwidth, clip, trim=40 40 40 40]{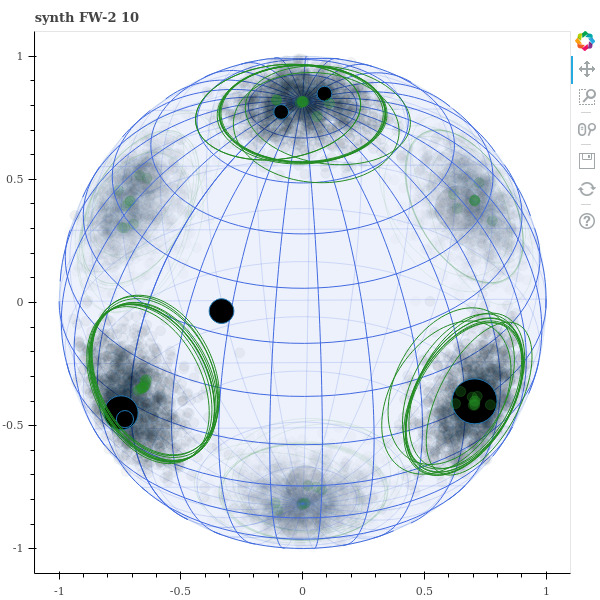}
\includegraphics[width=.32\columnwidth, clip, trim=40 40 40 40]{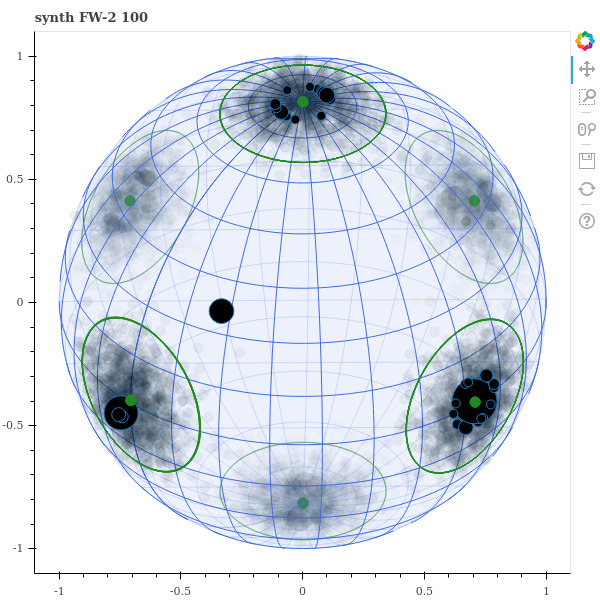}
\includegraphics[width=.32\columnwidth, clip, trim=40 40 40 40]{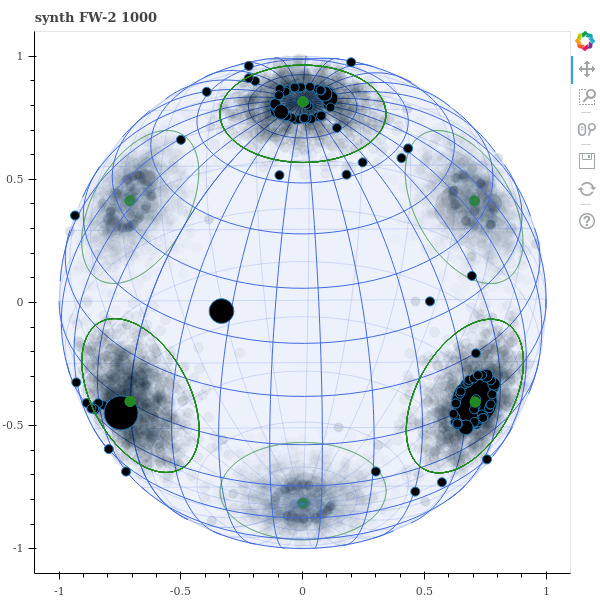}
\caption{Hilbert Frank--Wolfe}\label{fig:vmf_intuition_fw}
\end{subfigure}
\begin{subfigure}[t]{0.45\textwidth}
\includegraphics[width=\columnwidth]{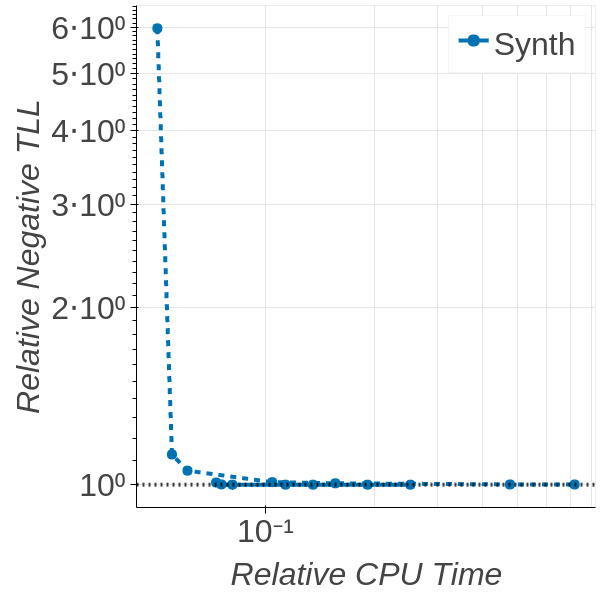}
\caption{}\label{fig:vmf_tll}
\end{subfigure}
\begin{subfigure}[t]{0.45\textwidth}
\includegraphics[width=\columnwidth]{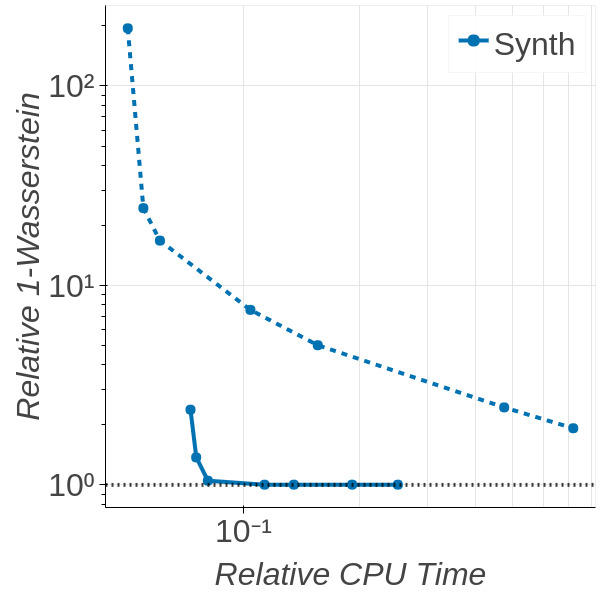}
\caption{}\label{fig:vmf_w1}
\end{subfigure}
\caption{(\ref{fig:vmf_intuition_rand}-\ref{fig:vmf_intuition_fw}): Comparison of different coreset constructions for directional clustering,
showing example coreset posterior mean clusters (green),
and a single trace of coreset construction (black) for $M =$ 10, 100, and 1,000. The radius of each coreset point indicates its weight.
(\ref{fig:vmf_tll}, \ref{fig:vmf_w1}): A comparison of negative test log-likelihood (\ref{fig:vmf_tll}) and 1-Wasserstein distance (\ref{fig:vmf_w1})
versus computation time for Frank--Wolfe (solid) and uniform random subsampling (dashed) on the directional clustering model.
Both axes are normalized using results from running MCMC on the full dataset; see \cref{sec:expt_methods}.
}\label{fig:vmf_expt}
\end{figure}

\begin{figure}[t!]
\begin{subfigure}[t]{0.32\textwidth}
\includegraphics[width=\columnwidth]{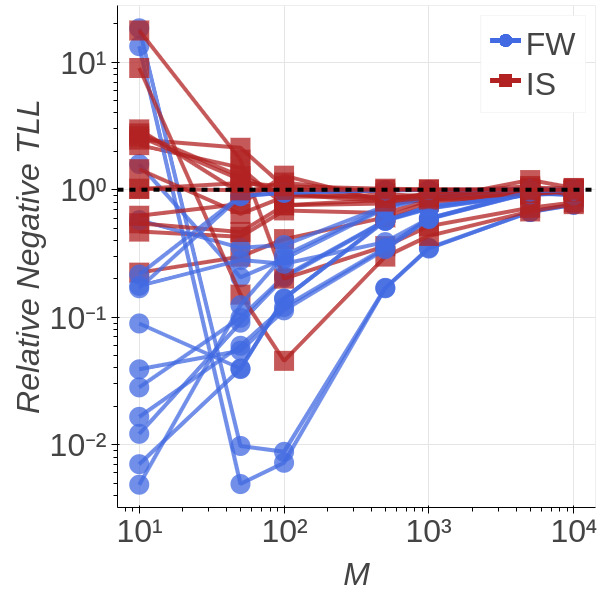}
\caption{}\label{fig:multi_tll}
\end{subfigure}
\begin{subfigure}[t]{0.32\textwidth}
\includegraphics[width=\columnwidth]{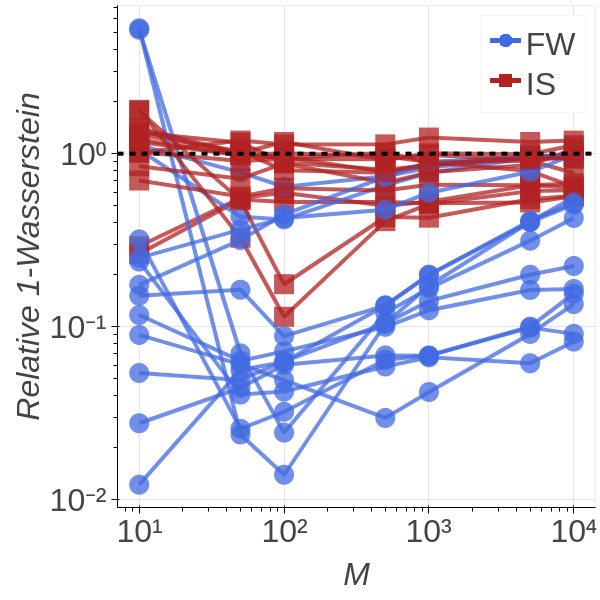}
\caption{}\label{fig:multi_w1}
\end{subfigure}
\begin{subfigure}[t]{0.32\textwidth}
\includegraphics[width=\columnwidth]{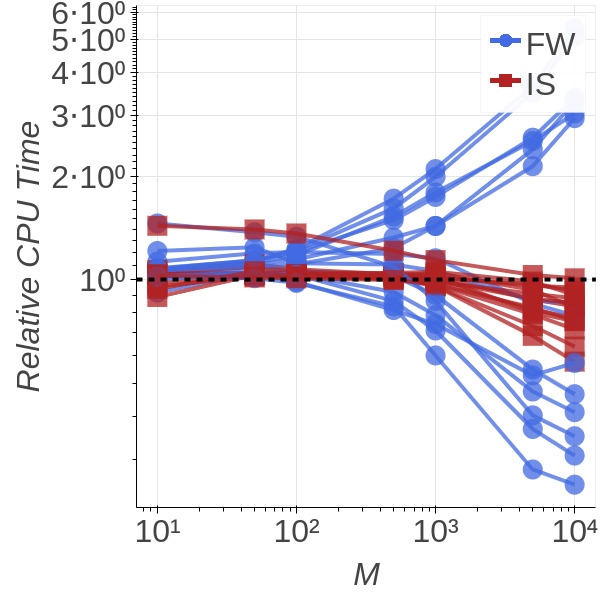}
\caption{}\label{fig:multi_t}
\end{subfigure}
\caption{A comparison of (\ref{fig:multi_tll}) negative test log-likelihood, (\ref{fig:multi_w1}) 1-Wasserstein posterior distance estimate,
and (\ref{fig:multi_t}) computation time  versus coreset construction iterations $M$ for Frank--Wolfe (blue), importance sampling (red), and uniformly random subsampling (dashed black) 
across all models and datasets. All metrics are normalized to the median value for uniformly random subsampling; see \cref{sec:expt_methods}.
}\label{fig:multi_quantitative}
\end{figure}

\cref{fig:lr_expt,fig:poiss_expt,fig:vmf_expt} show the experimental results for logistic regression, Poisson regression, and directional clustering, respectively.
The visual comparisons of coreset construction for all models mimic the results of the synthetic evaluation in \cref{fig:gauss_expt}.
For all the algorithms, the approximate posterior converges to the true posterior on the full dataset as more coreset points are added;
and the Frank--Wolfe Hilbert coreset construction selects the most useful points incrementally, creating intuitive coresets that outperform
all the other methods. For example,
in the logistic regression model, the uniform coreset construction sensitivities are based on the proximity of the 
data to the centers of a $K$-clustering, and do not directly incorporate information about the classification boundary. This construction therefore
generally favors sampling points on the periphery of the dataset and assigns high weight to those near the center.
The Hilbert importance sampling algorithm, in contrast, directly considers the logistic regression problem; it favors sampling points lying along the boundary 
and assigns high weight to points orthogonal to it, thereby fixing the boundary plane more accurately.
The Hilbert Frank--Wolfe algorithm selects a single point closely aligned with the classification boundary normal, and then
refines its estimate with points near the boundary. This enables it to use far fewer coreset points to achieve a more accurate posterior estimate than the sampling-based
methods. Similar statements hold for the two other models: in the Poisson regression model, the Hilbert Frank--Wolfe algorithm chooses a point closest to the true parameter and then
refines its estimate using far away points; and in the directional clustering model, the Hilbert Frank--Wolfe algorithm initially selects points near the cluster centers 
and then refines the estimates with points in each cluster far from its center.
The quantitative results demonstrate the strength of the Hilbert Frank--Wolfe coreset construction algorithm:
for a given computational time budget, this algorithm provides orders of magnitude reduction in both error metrics over uniformly random subsampling. 
In addition, the Hilbert Frank--Wolfe coreset achieves the same negative test log-likelihood as the full dataset in roughly a tenth of the computation time. 
These statements hold across all models and datasets considered. 

\cref{fig:multi_quantitative} provides a summary of the performance of Hilbert coresets as a function of construction iterations $M$ 
across all models and datasets. This demonstrates its power not only as a scalable inference method but also as a dataset compression technique.
For any value of $M$ and across a wide variety of models and datasets, Hilbert coresets (both importance sampling and Frank--Wolfe-based constructions) 
provide a significant improvement in posterior approximation quality over uniformly random subsampling, with comparable computation time. This figure also shows a rather surprising result:
not only does the Frank--Wolfe-based method provide improved posterior estimates, it also can sometimes have \emph{reduced overall computational cost} compared to
uniformly random subsampling for fixed $M$. This is due to the fact that $M$ is an upper bound on the coreset size; the Frank--Wolfe algorithm often selects the same
point multiple times, leading to coresets of size $\ll M$, whereas subsampling techniques always have coresets of size $\approx M$.
Since the cost of posterior inference scales with the coreset size and dominates the cost of setting up either coreset construction algorithm, the 
Hilbert Frank--Wolfe method has a reduced overall cost.
Generally speaking, we expect the Hilbert coreset methods to be slower than random subsampling for small $M$, where $\hat\pi$ setup and random projection dominates the time cost, but 
the Frank--Wolfe method to sometimes be faster for large $M$ where the smaller coreset provides a significant inferential boost.

Although more detailed results for Hilbert importance sampling, uniform coresets, and the weighted 2-norm from \cref{eq:norm2} are deferred to \cref{sec:additional_results},
we do provide a brief summary here. \cref{fig:is_quantitative} shows that Hilbert importance sampling provides comparable performance to both uniform coresets and uniformly random subsampling on 
all models and datasets. 
The Hilbert Frank--Wolfe coreset construction algorithm typically outperforms all random subsampling methods.
Finally, \cref{fig:isvis_quantitative,fig:fwvfw_quantitative} show that the weighted 2-norm and Fisher 
information norm perform similarly in all cases considered.

\section{Conclusion}
This paper presented a fully-automated, scalable, and theoretically-sound Bayesian inference framework based on Hilbert coresets.
The algorithms proposed in this work are simple to implement, reliably provide high-quality posterior approximation at a fraction of the cost
of running inference on the full dataset, and enable experts and nonexperts alike to conduct sophisticated modeling and exploratory analysis at scale.
There are many avenues for future work, including exploring the application of Hilbert coresets in more complex high-dimensional models,
using alternate Hilbert norms, connecting the norms proposed in the present work to more well-known measures of posterior discrepancy, 
investigating different choices of weighting function, obtaining tighter bounds on the quality of the random projection result,
and using variants of the Frank--Wolfe algorithm (e.g.~away-step, pairwise, and fully-corrective FW \citep{LacosteJulien15}) with stronger convergence guarantees.

\subsubsection*{Acknowledgments}
This research is supported by a Google Faculty Research Award,
an MIT Lincoln Laboratory Advanced Concepts Committee Award, and ONR grant
N00014-17-1-2072. We also thank Sushrutha Reddy for finding and correcting
a bug in the proof of Theorem 3.2.

\appendix
\section{Technical results and proofs}\label{sec:proofs}
In this section we provide proofs of the main results from the paper, along with
supporting technical lemmas. \cref{lem:azuma} is  
the martingale extension of Hoeffding's inequality \citep[Theorem 2.8, p.~34]{Boucheron13} known as Azuma's inequality.
\cref{lem:martingalebennet} is the martingale extension of Bennet's inequality \citep[Theorem 2.9, p.~35]{Boucheron13}.
\cref{lem:vectormomentbounds} provides bounds on the expectation and martingale differences of the norm of a vector constructed
by \iid sampling from a discrete distribution. Finally, \cref{lem:relint} is a geometric result
and \cref{lem:logisticbound} bounds iterates of the logistic equation, both of which are used in the proof of the Frank-Wolfe error bound.
\cref{lem:dnormratiovsddiam} provides a relationship between two vector alignment constants in the main text.
\bnlem[Azuma's Inequality]\label{lem:azuma}
Suppose $(Y_m)_{m=0}^M$ is a martingale adapted to the filtration $(\mcF_m)_{m=0}^M$.
If there is a constant $\xi$ such that for each $m\in\{1, \dots, M\}$,
\[
\left|Y_m - Y_{m-1}\right| &\leq \xi \quad \text{\as},
\]
then for all $\epsilon \geq 0$,
\[
\Pr\left(Y_M - Y_0 > \epsilon\right) &\leq e^{-\frac{\epsilon^2}{2 M \xi^2}}.
\]
\enlem

\bnlem[Martingale Bennet Inequality]\label{lem:martingalebennet}
Suppose $(Y_m)_{m=0}^M$ is a martingale adapted to the filtration $(\mcF_m)_{m=0}^M$.
If there are constants $\xi$ and $\tau^2$ such that for each $m\in \{1, \dots, M\}$, 
\[
\left|Y_m - Y_{m-1}\right| &\leq \xi  & \text{and}& & \EE\left[\left(Y_m-Y_{m-1}\right)^2 \given \mcF_{m-1}\right] &\leq \tau^2 \quad \text{\as}, 
\]
then for all $\epsilon \geq 0$,
\[
\Pr\left(Y_M - Y_0 > \epsilon\right) &\leq e^{-\frac{M\tau^2}{\xi^2} H\left(\frac{\epsilon\xi}{M\tau^2}\right)}, \quad H(x) \defined (1+x)\log(1+x) - x.
\]
\enlem

\bnlem\label{lem:vectormomentbounds}
Suppose $U$ and $\{U_m\}_{m=1}^M$ are \iid random vectors in a normed vector space with discrete support on $(u_n)_{n=1}^N$ with probabilities $(p_n)_{n=1}^N$, 
and
\[
Y &\defined \left\|\frac{1}{M}\sum_{m=1}^M U_m - \EE\left[U\right]\right\|. 
\]
Then we have the following results.
\benum
\item\label{lem:vmb_banach}Suppose $\dim\left(u_n\right)_{n=1}^N \leq d$ where $\dim$ is given by \cref{defn:dim},
$\alpha_{n\cdot}\in[-1, 1]^d$ are the coefficients used to approximate $u_n$ in \cref{defn:dim},
and $A_j$ is a random variable equal to $\alpha_{nj}$ when $U = u_n$.
Then
\[
\EE\left[Y\right] \leq \frac{d}{\sqrt{M}}\left(\sum_{n=1}^N\left\|u_n\right\|\sqrt{\frac{p_n(1-p_n)}{N}}
+ \sqrt{\var\left[\left\|UA_j\right\|\right]}\right).
\]
\item\label{lem:vmb_hilbert} If the norm is a Hilbert norm,
\[
\EE\left[Y\right] &\leq \frac{1}{\sqrt{M}}\sqrt{\EE\left[\left\|U\right\|^2\right] - \left\|\EE\left[U\right]\right\|^2}.
\]
\item\label{lem:vmb_martingale} The random variable $Y_m \defined \EE\left[Y \given \mcF_m \right]$ with $\mcF_m$ the $\sigma$-algebra generated by $U_1, \dots, U_m$ is a martingale that satisfies, for $m\geq 1$, both 
\[
\left| Y_m - Y_{m-1}\right| &\leq \frac{1}{M}\max_{n, \ell} \|u_n - u_\ell\|
\]
and
\[
\EE\left[\left(Y_m-Y_{m-1}\right)^2 \given \mcF_{m-1}\right] &\leq \frac{1}{M^2} \EE\left[\left\|U - U_1\right\|^2\right]
\]
almost surely.
\eenum
\enlem
\bprf
\benum
\item 
Using the triangle inequality, denoting the number of times vector $u_n$ is sampled as $M_n$,
\[
\EE\left[Y\right] \leq &\frac{1}{M}\sum_{n=1}^N\frac{d\left\|u_n\right\|}{\sqrt{N}}\EE\left[\left|M_n - Mp_n\right|\right] \nonumber \\ 
+ &\frac{1}{M}\sum_{j=1}^d\EE\left[\left| \sum_{n=1}^N \left(M_n-Mp_n\right)\|u_n\|\alpha_{nj}\right|\right].
\]
Bounding $\EE\left[\left|\cdot\right|\right] \leq \sqrt{\EE\left[(\cdot)^2\right]}$ via Jensen's inequality and evaluating the multinomial variances yields the desired result.
\item This follows from Jensen's inequality to write $\EE\left[Y\right] \leq \sqrt{\EE\left[Y^2\right]}$ and the expansion of the squared norm. 
\item $(Y_m)_{m=0}^M$ is a standard Doob martingale with $Y_0 = \EE\left[Y\right]$. Letting $U'_\ell = U_\ell$ for $\ell \neq m$ and $U'_m$ be an independent random variable with $U'_m \eqD U_m$,
by the triangle inequality we have
\[
&\left| Y_m - Y_{m-1}\right| \\
= &\left| \EE\left[Y \given \mcF_m\right] - Y_{m-1} \right|\\
= &\left|\EE\left[\left\|\frac{1}{M}\sum_{\ell=1}^M U_\ell - \EE\left[U\right]\right\| \given \mcF_m\right]  - Y_{m-1}\right|\\
= &\left|\EE\left[\left\|\frac{1}{M}\left(U_m - U'_m\right) + \frac{1}{M}\sum_{\ell=1}^M U'_\ell - \EE\left[U\right]\right\| \given \mcF_m\right]  - Y_{m-1}\right|\\
\leq &\left|\frac{1}{M}\EE\left[\left\|U_m - U'_m\right\| \given \mcF_m\right] + \EE\left[\left\|\frac{1}{M}\sum_{\ell=1}^M U'_\ell - \EE\left[U\right]\right\| \given \mcF_m\right]  - Y_{m-1}\right|\\
= &\frac{1}{M}\EE\left[\left\|U_m - U'_m\right\| \given \mcF_m\right]\label{eq:triintermediate}\\
\leq &\frac{1}{M}\max_{n, \ell} \|u_n - u_\ell\|.
\]
Next, using \cref{eq:triintermediate} and Jensen's inequality, we have that
\[
\EE\left[\left(Y_m-Y_{m-1}\right)^2 \given \mcF_{m-1}\right]  
& \leq\EE\left[\EE\left[\frac{1}{M}\left\|U_m - U'_m\right\|\given \mcF_m\right]^2 \given\mcF_{m-1}\right]\\
& \leq\frac{1}{M^2}\EE\left[\left\|U_m - U'_m\right\|^2\given \mcF_{m-1}\right]\\
& =\frac{1}{M^2}\EE\left[\left\|U_m - U'_m\right\|^2\right].
\]
\eenum
\eprf
\bprfof{\cref{thm:supremumimportancesampling}}
Set $\delta \in (0, 1)$. Rearranging the results of \cref{lem:azuma},
we have that with probability $\geq 1-\delta$,
\[
Y_M \leq Y_0 + \sqrt{2M\xi^2\log\frac{1}{\delta}}.
\]
We now apply the results of \cref{lem:vectormomentbounds}(\ref{lem:vmb_banach}) and \cref{lem:vectormomentbounds}(\ref{lem:vmb_martingale}), 
noting that $\sum_{n=1}^N\sqrt{p_n(1-p_n)}$ is maximized when $p_n = 1/N$,
where the discrete distribution 
is specified by atoms $u_n = \sigma \mcL_n/\sigma_n$ with probabilities $\sigma_n/\sigma$ for $n\in[N]$.
Further, when applying \cref{lem:vectormomentbounds}(\ref{lem:vmb_banach}), note that
\[
\var\|UA_j\| &= \sigma^2 \var|A_j|\\
&= \sigma^2 \var\left(|A_j|-\frac{1}{2}\right) \leq \sigma^2\EE\left[\left(|A_j|-\frac{1}{2}\right)^2\right] \leq \frac{\sigma^2}{4}.
\]
This yields
\[
Y_0 &\leq \frac{3\sigma d}{2\sqrt{M}} & \xi &= \frac{\sigma\ddiam}{M}.
\]
Substituting these results into the above expression,
\[
Y_M \leq \frac{\sigma}{\sqrt{M}}\left(\frac{3}{2}\dim\left(\mcL_n\right)_{n=1}^N + \ddiam\sqrt{2\log\frac{1}{\delta}}\right).
\]
\eprfof

\bprfof{\cref{thm:importancesampling}}
Set $\delta \in (0, 1)$. Rearranging the results of \cref{lem:azuma,lem:martingalebennet}, we have that
with probability $\geq 1-\delta$,
\[
Y_M \leq Y_0 + \min\left( \sqrt{2M\xi^2\log\frac{1}{\delta}}, \frac{M\tau^2}{\xi}H^{-1}\left(\frac{\xi^2}{M\tau^2}\log\frac{1}{\delta}\right)\right).
\]
We now apply the results of \cref{lem:vectormomentbounds}(\ref{lem:vmb_martingale}), where the discrete distribution 
is specified by atoms $u_n = \mcL_n/p_n$ with probabilities $p_n$ for $n\in[N]$.
Define $M_n$ to be the number of times index $n$ is sampled; then $(M_1, \dots, M_N) \dist \distMulti(M, (p_n)_{n=1}^N)$.
Then since our vectors are in a Hilbert space, we use \cref{lem:vectormomentbounds}(\ref{lem:vmb_hilbert}) and \cref{lem:vectormomentbounds}(\ref{lem:vmb_martingale}) to find that
\[
Y_M &= \left\|\frac{1}{M}\sum_{m=1}^M U_m - \EE\left[U\right]\right\|
= \left\|\sum_{n=1}^N \frac{M_n}{M p_n} \mcL_n - \mcL\right\|\\
Y_0 &\leq \sqrt{\frac{1}{M}\left(\sum_{n=1}^N \frac{\|\mcL_n\|^2}{p_n} - \left\|\mcL\right\|^2\right)}\\
\xi &= \frac{1}{M} \max_{m, n}\left\| \frac{\mcL_n}{p_n} - \frac{\mcL_m}{p_m}\right\|\\
\tau^2 &= \frac{1}{M^2}\EE\left[\left\|U_m-U'_m\right\|^2\right]
= \frac{2}{M^2}\left(\sum_{n=1}^N\frac{\|\mcL_n\|^2}{p_n} - \left\|\mcL\right\|^2\right).
\]
Minimizing both $\tau^2$ and $Y_0$ over $(p_n)_{n=1}^N$ by setting the derivative to 0 yields 
\[
p_n &= \frac{\|\mcL_n\|}{\sigma} & \sigma &\defined \sum_{n=1}^N \|\mcL_n\| .\label{eq:optimalsamplingprobs}
\]
Finally, we have that 
\[
Y_0 &\leq \sqrt{\frac{1}{M}\sigma^2\dnormratio^2} & 
\tau^2 &= \frac{2\sigma^2}{M^2}\left(1 - \frac{\|\mcL\|^2}{\sigma^2}\right) = \frac{2\sigma^2\dnormratio^2}{M^2} &
\xi &= \frac{\sigma}{M}\ddiam, \label{eq:expectationboundis}
\]
and $\frac{M_n}{M}\frac{1}{p_n} = W_n$ from \cref{alg:importancesampling}, so
\[
\left\|\mcL(W) - \mcL\right\| & \leq \frac{\sigma\dnormratio}{\sqrt{M}} +
\min\left( \sqrt{2\frac{\sigma^2\ddiam^2}{M}\log\frac{1}{\delta}}, \frac{2\sigma\dnormratio^2}{\ddiam}H^{-1}\left(\frac{\ddiam^2}{2M \dnormratio^2}\log\frac{1}{\delta}\right)\right)\\
& = \frac{\sigma}{\sqrt{M}}\left(\dnormratio + 
 \dM\sqrt{2 \log\frac{1}{\delta}}\right)\\
\dM &\defined \min\left( \ddiam,  \dnormratio\sqrt{\frac{2M\dnormratio^2}{\ddiam^2\log\frac{1}{\delta}}}H^{-1}\!\!\left(\frac{\ddiam^2\log\frac{1}{\delta}}{2M \dnormratio^2}\right)\right).
\]
\eprfof

\bnlem\label{lem:dnormratiovsddiam}
Given a Hilbert norm, $\ddiam$ from \cref{eq:sigddiamdefn} 
and $\dnormratio$ from \cref{eq:dnormratiodefn} satisfy
\[
\dnormratio &\leq \frac{\ddiam}{\sqrt{2}}.
\]
\enlem
\bprf
Noting $\left\|\mcL\right\|^2 = \left<\mcL, \mcL\right>$ and expanding the definition from \cref{eq:dnormratiodefn},
\[
\dnormratio^2 
= 1-\sum_{n, m = 1}^N\frac{\sigma_n\sigma_m}{\sigma^2} \left<\frac{\mcL_n}{\sigma_n}, \frac{\mcL_m}{\sigma_m}\right>
\leq 1-\min_{n, m \in [N]} \left<\frac{\mcL_n}{\sigma_n}, \frac{\mcL_m}{\sigma_m}\right> = \frac{1}{2}\ddiam^2
\]
where the inequality follows from $\sum_{n, m=1}^N \frac{\sigma_n\sigma_m}{\sigma^2} = 1$.
\eprf

\bnlem\label{lem:relint}
$\mcL$ is in the relative interior
of the convex hull of $\left\{\frac{\sigma}{\left\|\mcL_n\right\|}\mcL_n\right\}_{n=1}^N$.
\enlem
\bprf
First, since $K$ (the kernel matrix of inner products defined in \cref{eq:modifiedproblem}) is a symmetric postive semidefinite $N\times N$ matrix, there exists an $N\times N$ matrix $U$ such that
$\left\|\mcL(w)\right\|^2 = w^TKw = w^TU^TUw = \left\|u(w)\right\|_2^2$, where $u(w) \defined \sum_{n=1}^N w_nu_n$, $u\defined \sum_{n=1}^N u_n$, and $u_n\in\reals^N$ are the columns of $U$.
Therefore the mapping $\mcL(w) \to u(w)$ is a linear isometry from the Hilbert space to $\reals^N$, so if $u$ is in the relative interior
of the convex hull of $\left\{\frac{\sigma}{\left\|u_n\right\|}u_n\right\}_{n=1}^N$, the result follows.
Let $y$ be any other point in the convex hull in $\reals^N$, with coefficients $\gamma_n$.
If we set 
\[
\lambda = \min_{n : \gamma_n > \frac{\|u_n\|}{\sigma}} \frac{\gamma_n}{\gamma_n-\frac{\|u_n\|}{\sigma}}
\]
where the minimum of an empty set is defined to be $\infty$, 
then $\lambda u + (1-\lambda)y$ is in the convex hull and $\lambda > 1$.
Since for any point $y$ we can find such a $\lambda$, the result follows from \citep[Theorem 6.4, p.~47]{Rockafeller70}.
\eprf

\bnlem\label{lem:logisticbound}
The logistic recursion,
\[
x_{n+1} \leq \alpha x_n(1-x_n),
\]
for $x_0, \alpha \in [0, 1]$ satisfies
\[
\forall n \in \nats, \quad x_n \leq \frac{x_0}{\alpha^{-n} + x_0n}.
\]
\enlem
\bprf
The proof proceeds by induction. The bound holds at $n=0$ since
\[
x_0 &\leq \frac{x_0}{\alpha^0 + 0} = x_0.
\]
Since $1-x \leq 1/(1+x)$, for any $n\geq 0$,
\[
x_{n+1} &\leq \alpha x_n(1-x_n) \leq \alpha \frac{x_n}{1+x_n}.
\]
Assuming the bound holds for any $n\geq 0$,
and noting $x/(1+x)$ is monotone increasing for $x\geq 0$, we
can substitute the bound yielding
\[
x_{n+1} &\leq \alpha \frac{\frac{x_0}{\alpha^{-n}+nx_0}}{1+\frac{x_0}{\alpha^{-n}+nx_0}} = \alpha \frac{x_0}{\alpha^{-n}+nx_0+x_0} = \frac{x_0}{\alpha^{-(n+1)}+\alpha^{-1}(n+1)x_0}.
\]
The final result follows since $\alpha^{-1} \geq 1$.
\eprf

\bprfof{\cref{lem:linesearch}}
Let $w_t$ be the weight vector at iteration $t$ in \cref{alg:frankwolfe}, 
and let $f_t$ and $d_t$ be the Frank-Wolfe vertex index and direction, respectively, from \cref{eq:fwdir}.
For brevity, denote the cost $J(w) \defined (w - 1)^TK(w-1)$.
For any $\gamma \in \reals$, if we let $w_{t+1} = w_t + \gamma d_t$
we have that
\[
J(w_{t+1}) =J(w_t) + 2\gamma d_t^TK(w_t - 1) + \gamma^2d_t^TKd_t.\label{eq:lineeq}
\]
Minimizing \cref{eq:lineeq} over $\gamma\in\reals$ yields \cref{eq:linesearch} (expressed as a quadratic form with gram matrix $K$),
\[
\gamma_t &= \frac{d_t^TK(1-w_t)}{d_t^TKd_t}. \label{eq:linesearchKform}
\]
Suppose $\gamma_t < 0$. Then $d_t^TK(1-w_t) < 0$; but $d_t$ maximizes this product over feasible directions, so 
\[
0 > d_t^TK(1-w_t) > (1-w_t)^TK(1-w_t) = J(w_t) \geq 0, \label{eq:fwmtJ}
\]
which is a contradiction. Now suppose $\gamma_t > 1$. Then
\[
d_t^TK(1-w_t) > d_t^TKd_t, 
\]
and \cref{eq:fwmtJ} holds again,
so if we were to select $\gamma = 1$ in \cref{eq:lineeq}, we would have
\[
0 \leq J(w_{t+1}) &< J(w_t) + d_t^TK(w_t - 1)\leq 0, 
\]
which is another contradiction, so $\gamma_t \leq 1$. 
Therefore $\gamma_t \in [0, 1]$
\eprfof
\bprfof{\cref{thm:frankwolfe}}
Using the same notation as the proof of \cref{lem:linesearch} above,
first note that $J(w_0) \leq \sigma^2 \dnormratio^2$ as initialized by \cref{eq:fwinit}: 
for any $\xi \in\reals_+^N$ with $\sum_n \xi_n = 1$,
\[
\frac{J(w_0)}{\sigma^2} &= 
1
- 2\left<\frac{\mcL_{f_0}}{\sigma_{f_0}}, \frac{\mcL}{\sigma}\right>
+ \frac{\left\|\mcL\right\|^2}{\sigma^2}
\leq
1
- 2\sum_{n=1}^N \xi_n \left<\frac{\mcL_{n}}{\sigma_{n}}, \frac{\mcL}{\sigma}\right>
+ \frac{\left\|\mcL\right\|^2}{\sigma^2}
\]
since $f_0$ maximizes $\left<\mcL, \mcL_n/\sigma_n\right>$ over $n\in[N]$,
and picking $\xi_n = \sigma_n/\sigma$ yields
\[
\frac{J(w_0)}{\sigma^2}
\leq
1
- 2\sum_{n=1}^N \left<\frac{\mcL_{n}}{\sigma}, \frac{\mcL}{\sigma}\right>
+ \frac{\left\|\mcL\right\|^2}{\sigma^2}
=
1 - \frac{\left\|\mcL\right\|^2}{\sigma^2} 
= \dnormratio^2.
\]
By \cref{lem:linesearch}, we are guaranteed that each Frank-Wolfe iterate using exact line search is feasible,
and substituting \cref{eq:linesearchKform} into \cref{eq:lineeq} yields
\[
J(w_{t+1}) &= J(w_t) - \frac{\left(d_t^TK(1-w_t)\right)^2}{d_t^TKd_t}\\
&=J(w_t) \left( 1 - \left<\frac{\frac{\sigma}{\sigma_{f_t}}\mcL_{f_t} - \mcL(w_t)}{\|\frac{\sigma}{\sigma_{f_t}}\mcL_{f_t}-\mcL(w_t)\|}, \frac{\mcL - \mcL(w_t)}{\|\mcL-\mcL(w_t)\|}\right>^2\right). \label{eq:linesearchexact}
\]
We now employ a technique due to \citet{Guelat86}: 
by \cref{lem:relint}, $\mcL$ is in the relative interior of the convex hull of the $\left\{\frac{\sigma}{\sigma_n} \mcL_n\right\}_{n=1}^N$,
so there exists an $r > 0$ such that for any feasible $w$,
\[
\mcL(w) + (\|\mcL - \mcL(w)\| + r)\frac{\mcL-\mcL(w)}{\|\mcL-\mcL(w)\|}
\]
is also in the convex hull. Thus, since the Frank-Wolfe vertex $\frac{\sigma}{\sigma_{f_t}}\mcL_{f_t}$ maximizes
$\left<\mcL(w) - \mcL(w_t), \mcL-\mcL(w_t)\right>$ over feasible $w$, we have that
\[
\left<\frac{\frac{\sigma}{\sigma_{f_t}}\mcL_{f_t} - \mcL(w_t)}{\|\frac{\sigma}{\sigma_{f_t}}\mcL_{f_t}-\mcL(w_t)\|}, \frac{\mcL - \mcL(w_t)}{\|\mcL-\mcL(w_t)\|}\right>
&\geq 
\left<\frac{(\|\mcL - \mcL(w_t)\| + r)\frac{\mcL-\mcL(w_t)}{\|\mcL-\mcL(w_t)\|}}{\|\frac{\sigma}{\sigma_{f_t}}\mcL_{f_t}-\mcL(w_t)\|}, \frac{\mcL - \mcL(w_t)}{\|\mcL-\mcL(w_t)\|}\right>\\
&= 
\frac{\sqrt{J(w_t)} + r}{\|\frac{\sigma}{\sigma_{f_t}}\mcL_{f_t}-\mcL(w_t)\|}\\
&\geq \frac{\sqrt{J(w_t)} + r}{\sigma \ddiam}.
\]
Substituting this into \cref{eq:linesearchexact} yields 
\[
J(w_{t+1}) &\leq J(w_t)\left(1 - \left(\frac{\sqrt{J(w_t)} + r}{\sigma \ddiam}\right)^2\right)
\leq J(w_t)\left(\doptwidth^2 - \frac{J(w_t)}{\sigma^2\ddiam^2}\right),
\]
where $\doptwidth \defined 1-\frac{r^2}{\sigma^2\ddiam^2}$. Defining $x_t \defined \frac{J(w_t)}{\sigma^2\ddiam^2\doptwidth^2}$, 
we have that $0 \leq x_t \leq 1$ and 
\[
x_{t+1} &\leq \doptwidth^2 x_t(1-x_t),
\]
and so \cref{lem:logisticbound} implies that
\[
\frac{J(w_t)}{\sigma^2\ddiam^2\doptwidth^2} &\leq \frac{ \frac{J(w_0)}{\sigma^2\ddiam^2\doptwidth^2}}{\doptwidth^{-2t} + \frac{J(w_0)}{\sigma^2\ddiam^2\doptwidth^2} t}.
\]
Further, since the function $\frac{a}{a+b}$ is monotonically increasing in $a$ for all $a, b \geq 0$, we can use the bound on the initial objective $J(w_0)$, yielding
\[
J(w_t) &\leq \frac{\sigma^2\dnormratio^2\ddiam^2\doptwidth^2}{\ddiam^2\doptwidth^{2-2t} + \dnormratio^2 t} 
\]
The proof concludes by noting that we compute $M-1$ iterations after initialization to construct a coreset of size $\leq M$.
The second stated bound results from the fact that $\ddiam \geq \dnormratio$ and $\doptwidth \leq 1$.

The weaker bound in the note after the theorem is a result of a technique 
very similar to that commonly found in past work \citep{Clarkson10,Jaggi13}: 
starting from \cref{eq:lineeq},
we bound $d_t^TKd_t \leq \sigma^2\ddiam^2$ and $d_t^TK(w_t-1) \leq -J(w_t)$,
and then use recursion to prove that $J(w_t) \leq \frac{4\sigma^2\ddiam^2}{3t+4}$
given $\gamma_t = \frac{2}{3t+4}$.
\eprfof

\bprfof{\cref{thm:bayesfw}}
Suppose $\max_{m, n}\left|\left<\mcL_n, \mcL_m\right> - v_n^Tv_m\right| \leq \epsilon$. Then
\[
(w-1)^TK(w-1) &- (w-1)^TV(w-1) \leq \notag \\ 
&\sum_{m,n} \left|w_n-1\right|\left|w_m-1\right|\left|\left<\mcL_n, \mcL_m\right> - v_n^Tv_m\right|
\leq \left\|w-1\right\|_1^2 \epsilon.
\]
We now bound the probability that the above inequality holds, assuming $D\grad\mcL_n(\mu)_d\grad\mcL_m(\mu)_d$ (when using $\mcD_{\hat\pi, F}$) or $\mcL_n(\mu)\mcL_m(\mu)$ (when using $\mcD_{\hat\pi, 2}$) is sub-Gaussian with constant $\xi^2$.
For brevity denote the true vector $\mcL_n$ as $\mcL_n$ and its random projection as $v_n$. Then
\[
&\Pr\left(\max_{m, n}\left|\left<\mcL_n, \mcL_m\right> - v_n^Tv_m\right| \geq \epsilon\right) \notag\\
\leq &\sum_{m,n}\Pr\left(\left|\left<\mcL_n, \mcL_m\right> - v_n^Tv_m\right| \geq \epsilon\right)\\
 \leq& N^2 \max_{m,n}\Pr\left(\left|\left<\mcL_n, \mcL_m\right> - v_n^Tv_m\right| \geq \epsilon\right)\\
=& N^2 \max_{m,n}\Pr\left(\left|\left<\mcL_n, \mcL_m\right> - \frac{1}{J}\sum_{j=1}^Jv_{nj}v_{mj}\right| \geq \epsilon\right)\\
\leq& 2 N^2  e^{-\frac{J\epsilon^2}{2 \xi^2}},
\]
using Hoeffding's inequality for sub-Gaussian variables. Thus if we fix $\delta \in (0, 1)$, with probability $\geq 1-\delta$,
\[
\sqrt{\frac{2\xi^2}{J}\log\frac{2N^2}{\delta}} \geq  \epsilon.
\]
Therefore, with probability $\geq 1-\delta$,
\[
\|\mcL(w)-\mcL\|^2 &\leq \|v(w)-v\|^2 + \|w-1\|_1^2\sqrt{\frac{2\xi^2}{J}\log\frac{2N^2}{\delta}}.
\]
\eprfof

\section{Additional Results}\label{sec:additional_results}
\begin{figure}[h!]
\begin{subfigure}[t]{0.45\textwidth}
\includegraphics[width=\columnwidth]{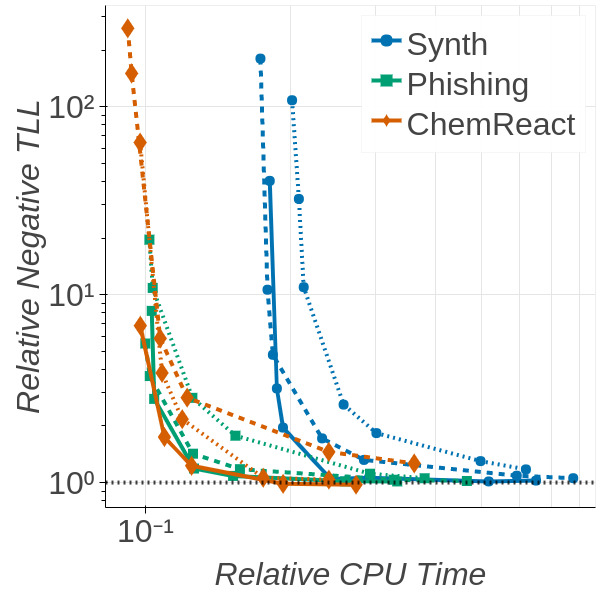}
\caption{}\label{fig:lr_is_tll}
\end{subfigure}
\begin{subfigure}[t]{0.45\textwidth}
\includegraphics[width=\columnwidth]{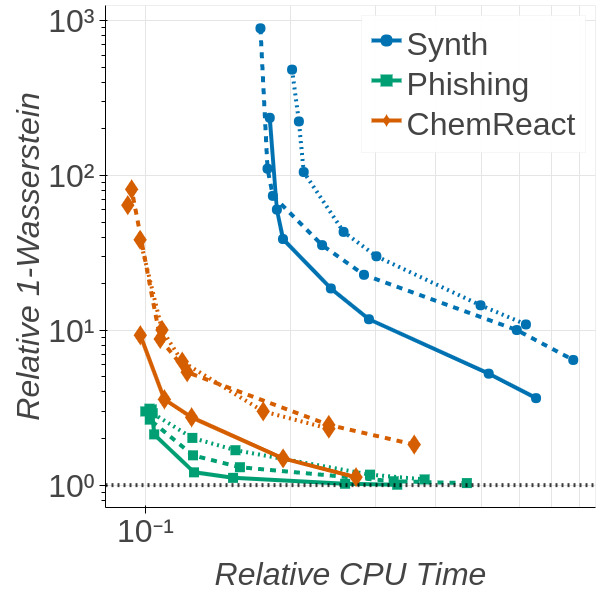}
\caption{}\label{fig:lr_is_w1}
\end{subfigure}
\begin{subfigure}[t]{0.45\textwidth}
\includegraphics[width=\columnwidth]{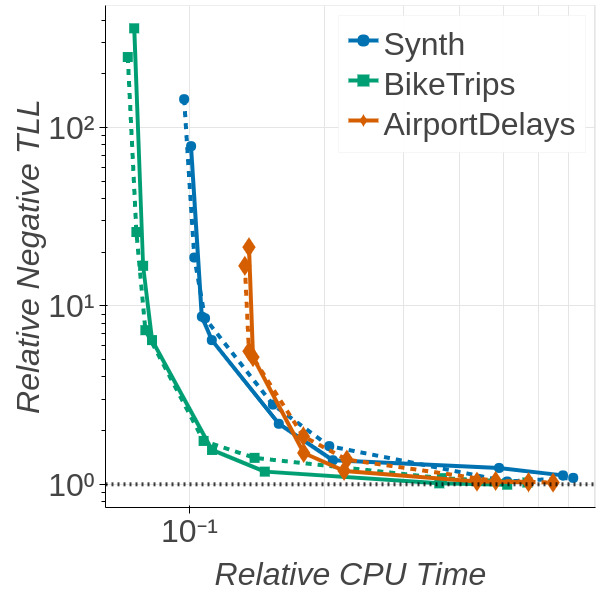}
\caption{}\label{fig:poiss_is_tll}
\end{subfigure}
\begin{subfigure}[t]{0.45\textwidth}
\includegraphics[width=\columnwidth]{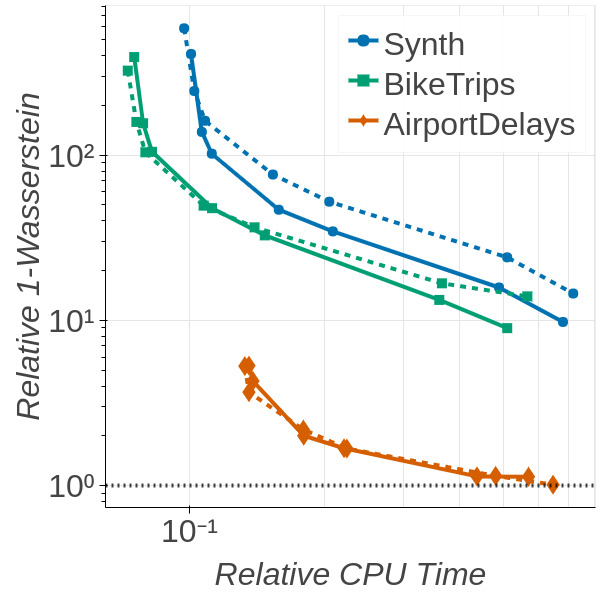}
\caption{}\label{fig:poiss_is_w1}
\end{subfigure}
\begin{subfigure}[t]{0.45\textwidth}
\includegraphics[width=\columnwidth]{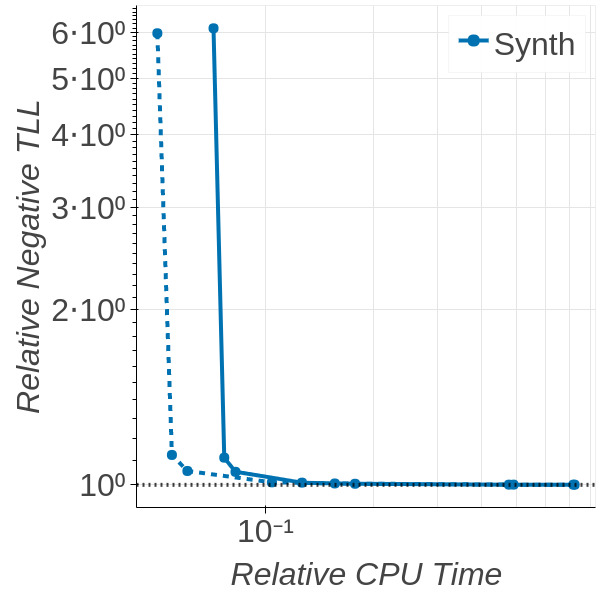}
\caption{}\label{fig:vmf_is_tll}
\end{subfigure}
\begin{subfigure}[t]{0.45\textwidth}
\includegraphics[width=\columnwidth]{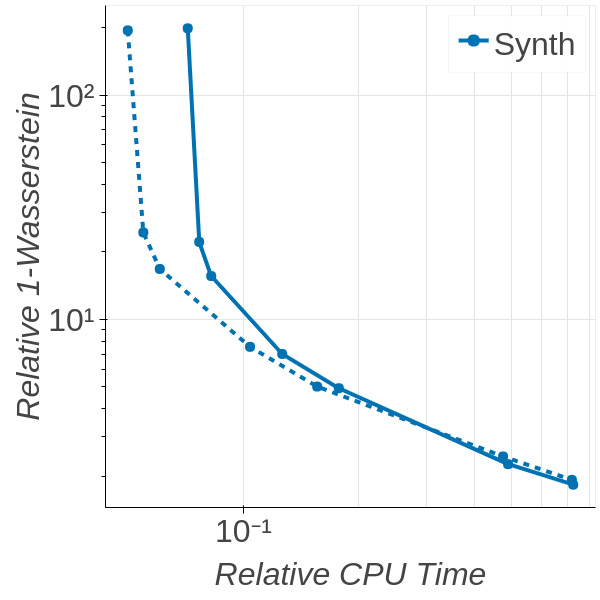}
\caption{}\label{fig:vmf_is_w1}
\end{subfigure}
\caption{Comparisons for IS-F (solid), uniform coresets (dotted), and uniform random subsampling (dashed) on 
(\ref{fig:lr_is_tll}, \ref{fig:lr_is_w1}) logistic regression,
(\ref{fig:poiss_is_tll}, \ref{fig:poiss_is_w1}) Poisson regression, and
(\ref{fig:vmf_is_tll}, \ref{fig:vmf_is_w1}) directional clustering.
Both axes are normalized; see \cref{sec:expt_methods}.
}\label{fig:is_quantitative}
\end{figure}

\begin{figure}[h!]
\begin{subfigure}[t]{0.45\textwidth}
\includegraphics[width=\columnwidth]{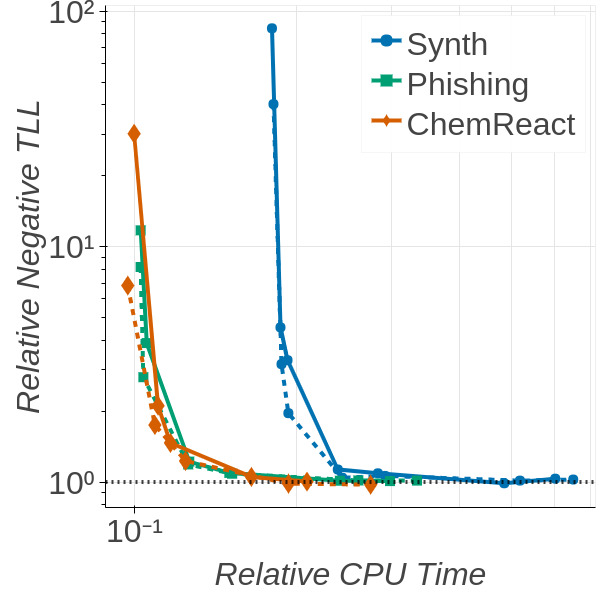}
\caption{}\label{fig:lr_isvis_tll}
\end{subfigure}
\begin{subfigure}[t]{0.45\textwidth}
\includegraphics[width=\columnwidth]{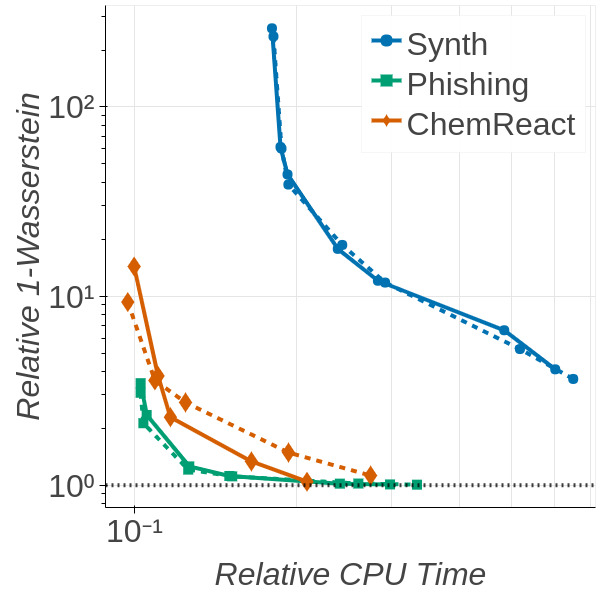}
\caption{}\label{fig:lr_isvis_w1}
\end{subfigure}
\begin{subfigure}[t]{0.45\textwidth}
\includegraphics[width=\columnwidth]{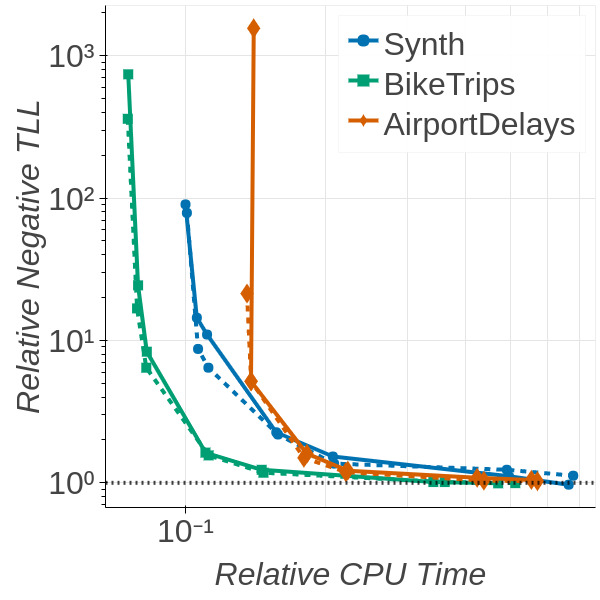}
\caption{}\label{fig:poiss_isvis_tll}
\end{subfigure}
\begin{subfigure}[t]{0.45\textwidth}
\includegraphics[width=\columnwidth]{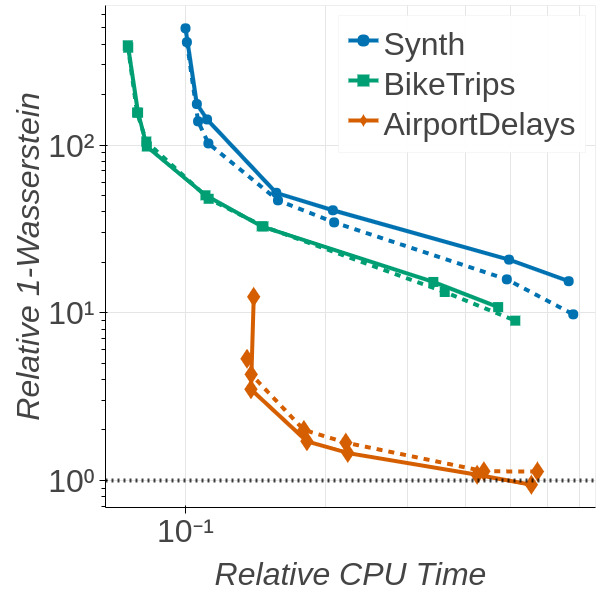}
\caption{}\label{fig:poiss_isvis_w1}
\end{subfigure}
\begin{subfigure}[t]{0.45\textwidth}
\includegraphics[width=\columnwidth]{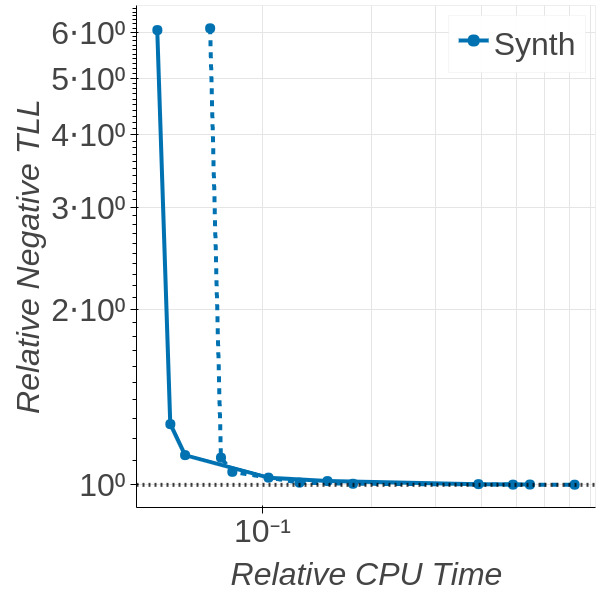}
\caption{}\label{fig:vmf_isvis_tll}
\end{subfigure}
\begin{subfigure}[t]{0.45\textwidth}
\includegraphics[width=\columnwidth]{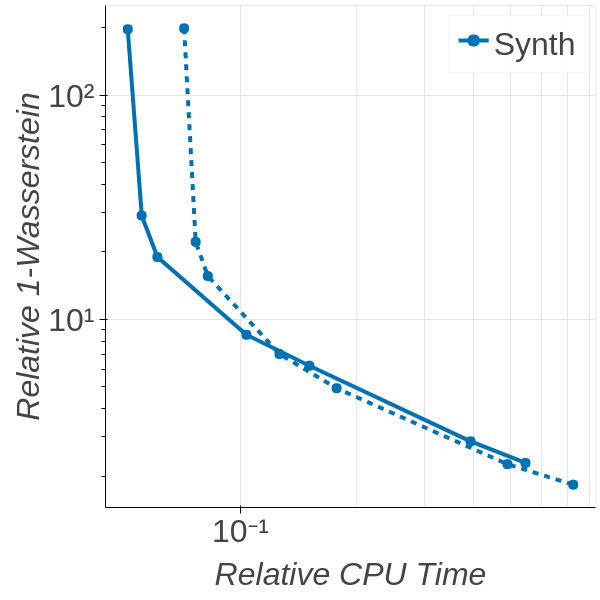}
\caption{}\label{fig:vmf_isvis_w1}
\end{subfigure}
\caption{Comparisons for IS-2 (solid) and IS-F (dashed) on 
(\ref{fig:lr_isvis_tll}, \ref{fig:lr_isvis_w1}) logistic regression,
(\ref{fig:poiss_isvis_tll}, \ref{fig:poiss_isvis_w1}) Poisson regression, and
(\ref{fig:vmf_isvis_tll}, \ref{fig:vmf_isvis_w1}) directional clustering.
Both axes are normalized; see \cref{sec:expt_methods}.
}\label{fig:isvis_quantitative}
\end{figure}

\begin{figure}[h!]
\begin{subfigure}[t]{0.45\textwidth}
\includegraphics[width=\columnwidth]{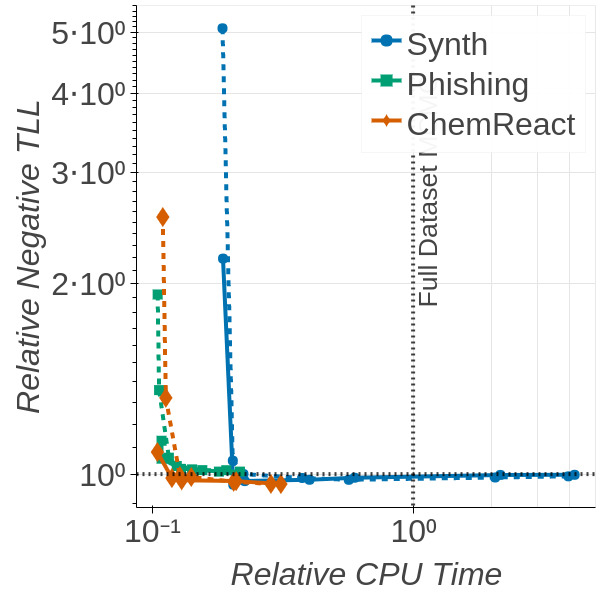}
\caption{}\label{fig:lr_fwvfw_tll}
\end{subfigure}
\begin{subfigure}[t]{0.45\textwidth}
\includegraphics[width=\columnwidth]{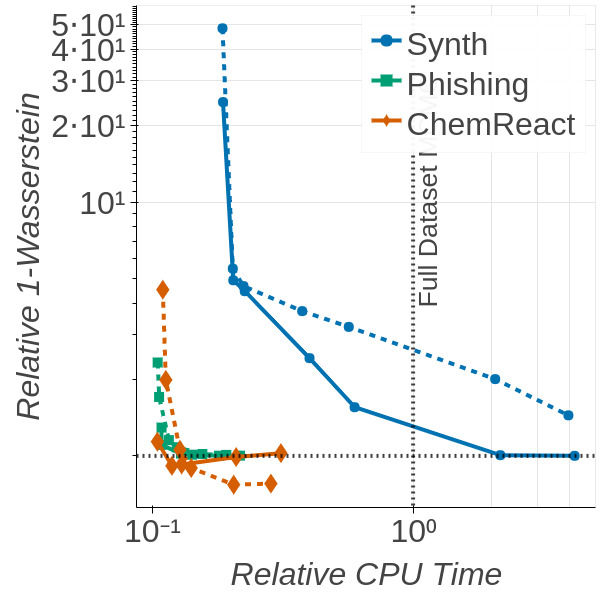}
\caption{}\label{fig:lr_fwvfw_w1}
\end{subfigure}
\begin{subfigure}[t]{0.45\textwidth}
\includegraphics[width=\columnwidth]{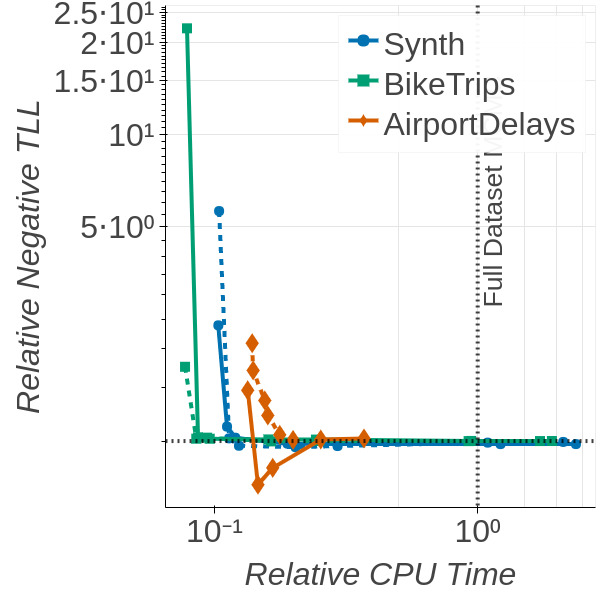}
\caption{}\label{fig:poiss_fwvfw_tll}
\end{subfigure}
\begin{subfigure}[t]{0.45\textwidth}
\includegraphics[width=\columnwidth]{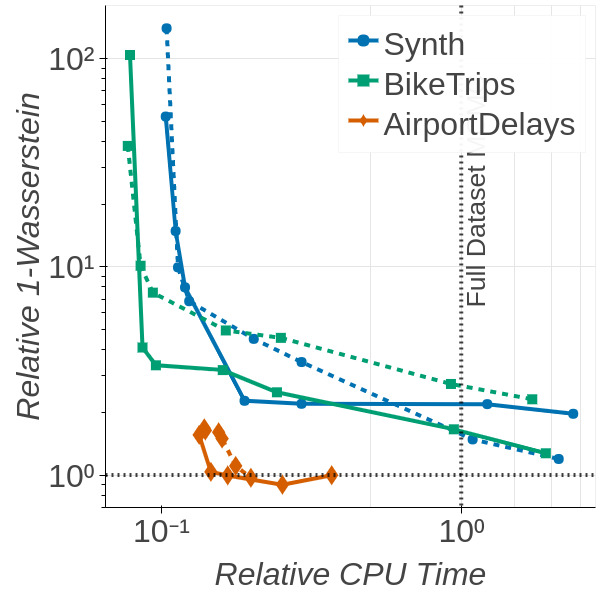}
\caption{}\label{fig:poiss_fwvfw_w1}
\end{subfigure}
\begin{subfigure}[t]{0.45\textwidth}
\includegraphics[width=\columnwidth]{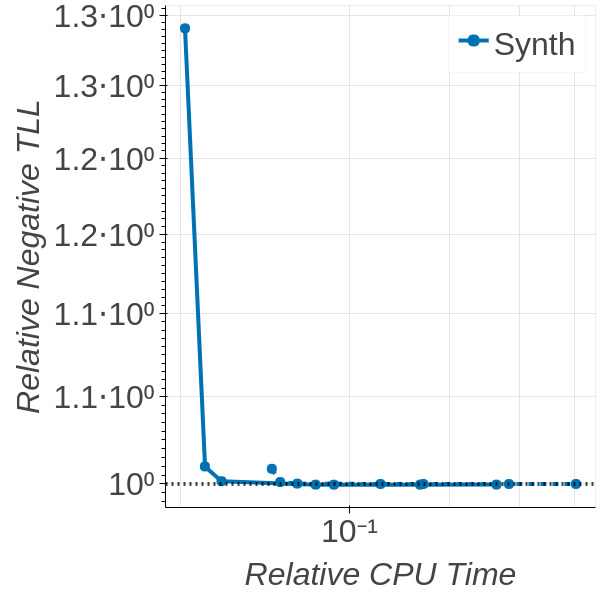}
\caption{}\label{fig:vmf_fwvfw_tll}
\end{subfigure}
\begin{subfigure}[t]{0.45\textwidth}
\includegraphics[width=\columnwidth]{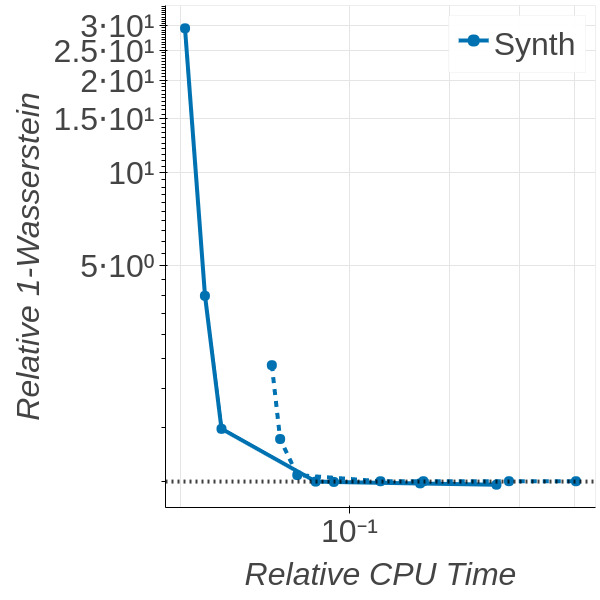}
\caption{}\label{fig:vmf_fwvfw_w1}
\end{subfigure}
\caption{Comparisons for FW-2 (solid) and FW-F (dashed) on 
(\ref{fig:lr_fwvfw_tll}, \ref{fig:lr_fwvfw_w1}) logistic regression,
(\ref{fig:poiss_fwvfw_tll}, \ref{fig:poiss_fwvfw_w1}) Poisson regression, and
(\ref{fig:vmf_fwvfw_tll}, \ref{fig:vmf_fwvfw_w1}) directional clustering.
Both axes are normalized; see \cref{sec:expt_methods}.
}\label{fig:fwvfw_quantitative}
\end{figure}

This section contains supplementary quantitative evaluations.
\cref{fig:is_quantitative} compares Hilbert importance sampling to
uniformly random subsampling and uniform coresets. These results
demonstrate that all subsampling techniques perform similarly,
with Hilbert coresets often the best choice of the three.
The Frank--Wolfe constructions outperform subsampling techniques
across all models and datasets considered.
\cref{fig:isvis_quantitative,fig:fwvfw_quantitative}
compare the weighted 2-norm from \cref{eq:norm2} to the weighted Fisher information norm from \cref{eq:normF}
in both importance sampling and Frank--Wolfe-based Hilbert coreset constructions.
These results show that the 2-norm and F-norm perform similarly in all cases.


\clearpage

\clearpage

\clearpage

\clearpage

\section{Derivation of the Gaussian uniform coreset sensitivity}\label{sec:normalderiv}
The sensitivity of observation $y_n$ used in the construction of a Bayesian coreset \citep{Huggins16} 
(ignoring constants) is
\[
N \sigma_n = \sup_{\mu\in\reals^d} \frac{N\mcL_n(\mu)}{\mcL(\mu)} &= \sup_{\mu\in\reals^d} 
\frac
{N\left(y_n-\mu\right)^T\left(y_n-\mu\right)}
{\sum_{m=1}^N\left(y_m-\mu\right)^T\left(y_m-\mu\right) }.
\]
By noting that
\[
\frac{1}{N}\sum_{m=1}^N\left(y_m-\mu\right)^T\left(y_m-\mu\right) 
&= \frac{1}{N}\sum_{m=1}^N y_m^Ty_m -\by^T\by + \left(\mu-\by\right)^T\left(\mu-\by\right),
\]
where $\by \defined \frac{1}{N}\sum_{m=1}^N y_m$,
we can keep the denominator constant by varying $\mu$ on the ball centered at $\by$ of constant radius.
The maximum of the numerator while keeping the denominator constant happens when $\mu$ lies on the 1d affine space
between $\by$ and $y_n$; so we can reparametrize $\mu = \lambda \by + (1-\lambda)y_n$ for $\lambda\in\reals$, yielding the optimization
\[
\sup_{\mu\in\reals^d} \frac{N\mcL_n(\mu)}{\mcL(\mu)} 
&= \sup_{\lambda\in\reals}
\frac
{\lambda^2\left(y_n-\by\right)^T\left(y_n-\by\right)}
{\frac{1}{N}\sum_{m=1}^N y_m^Ty_m -\by^T\by + (1-\lambda)^2\left(y_n-\by\right)^T\left(y_n-\by\right)}
\]
for which the optimum occurs at $\lambda^\star = \left(\frac{\frac{1}{N}\sum_{m=1}^N y_m^Ty_m -\by^T\by}{\left(y_n-\by\right)^T\left(y_n-\by\right)}+1\right)$ with value
\[
N\sigma_n = \sup_{\mu\in\reals^d} \frac{N\mcL_n(\mu)}{\mcL(\mu)}  
&= 1 + \frac{\left(y_n-\by\right)^T\left(y_n-\by\right)}{\frac{1}{N}\sum_{m=1}^Ny_m^Ty_m - \by^T\by}.
\]

\newpage
\small
\bibliographystyle{ba}
\bibliography{main}

\end{document}